\documentclass[10pt,journal,compsoc]{IEEEtran}

\ifCLASSOPTIONcompsoc
  \usepackage[nocompress]{cite}
\else
  \usepackage{cite}
\fi
\usepackage{amsmath,amsfonts}
\usepackage{algorithm}
\usepackage{array}
\usepackage{textcomp}
\usepackage{stfloats}
\usepackage{url}
\usepackage{verbatim}
\usepackage{graphicx}
\usepackage{cite}
\usepackage{amsmath,amssymb,amsfonts,bm,bbm}
\usepackage{multirow}
\usepackage{subfigure}
\usepackage{caption}
\usepackage{balance}
\usepackage{color}
\usepackage[hidelinks]{hyperref}
\makeatletter  
\newif\if@restonecol  
\makeatother

\usepackage[linesnumbered,ruled,vlined,algo2e]{algorithm2e}
\usepackage{algpseudocode}  
\usepackage{amsmath}  

\ifCLASSINFOpdf
\else
\fi
\begin{document}
%
\title{Mitigating Noisy Supervision Using Synthetic Samples with Soft Labels}
%
%
%
%

\author{Yangdi Lu,
        Wenbo He,
\IEEEcompsocitemizethanks{\IEEEcompsocthanksitem Y. Lu and W. He are with the Department of Computing and Software, McMaster University, Hamilton, Ontario L8S 4L7, Canada.\protect\\
E-mail: luy100@mcmaster.ca, hew11@mcmaster.ca }
\thanks{Manuscript received May 6, 2022.}}

\IEEEtitleabstractindextext{%
\begin{abstract}
Noisy labels are ubiquitous in real-world datasets, especially in the large-scale ones derived from crowdsourcing and web searching. It is challenging to train deep neural networks with noisy datasets since the networks are prone to overfitting the noisy labels during training, resulting in poor generalization performance. During a ``early learning'' phase, deep neural networks have been observed to fit the clean samples before memorizing the mislabeled samples. In this paper, we dig deeper into the representation distributions in the early learning phase and find that, regardless of their noisy labels, learned representations of images from the same category still congregate together. Inspired by it, we propose a framework that trains the model with new synthetic samples to mitigate the impact of noisy labels. Specifically, we propose a mixing strategy to create the synthetic samples by aggregating original samples with their top-$K$ nearest neighbours, wherein the weights are calculated using a mixture model learning from the per-sample loss distribution. To enhance the performance in the presence of extreme label noise, we estimate the soft targets by gradually correcting the noisy labels. Furthermore, we demonstrate that the estimated soft targets yield a more accurate approximation to ground truth labels and the proposed method produces a superior quality of learned representations with more separated and clearly bounded clusters. The extensive experiments in two benchmarks (CIFAR-10 and CIFAR-100) and two larg-scale real-world datasets (Clothing1M and Webvision) demonstrate that our approach outperforms the state-of-the-art methods and robustness of the learned representation. The code is available at {\color{magenta}\href{https://github.com/MacLLL/MixNN}{https://github.com/MacLLL/MixNN}}.


\end{abstract}

}

\maketitle

\IEEEdisplaynontitleabstractindextext

%
\IEEEpeerreviewmaketitle

\IEEEraisesectionheading{\section{Introduction}\label{sec:introduction}}

\IEEEPARstart{D}{eep} neural networks (DNNs) have achieved remarkable performance in a variety of applications (e.g. image classification and object detection). Despite the use of novel network architectures and efficient optimization algorithms, large-scale training data with correct labels is always required for these supervised tasks. However, obtaining such high-quality training data with human-annotated labels is extremely expensive and time-consuming in practice. Some non-expert sources, such as Amazon's Mechanical Turk and online searching engines, have been widely used to lower the high labeling cost. However, due to the limited knowledge and inadvertent mistakes, crowdsourced annotators cannot annotate specific tasks with 100\% accuracy, resulting in introducing \emph{noisy labels}. Even the most celebrated and highly-curated datasets, such as ImageNet \cite{deng2009imagenet}, are famously containing noisy labels. 

Unlike traditional supervised learning, which assumes that all label information is correct, we consider the training data contains a certain percentage of samples with incorrect labels. Training DNNs on such unreliable datasets is known to be highly affected as the significant number of model parameters render DNNs even overfit to noisy labels \cite{li2020gradient}. Zhang et al. \cite{zhang2018understanding} have empirically demonstrated that DNNs can easily fit an entire training dataset with any percentage of corrupted labels, and result in poor generalization capacity on a clean test set. Zhu et al. \cite{zhu2004class} have observed that the performance drop caused by label noise is more substantial than by other noises, such as feature noise. Therefore, it is crucial to develop learning algorithms that achieve superior generalization capability in the presence of noisy labels.

\begin{figure}[t]
	\begin{center}
		\includegraphics[width=0.8\linewidth]{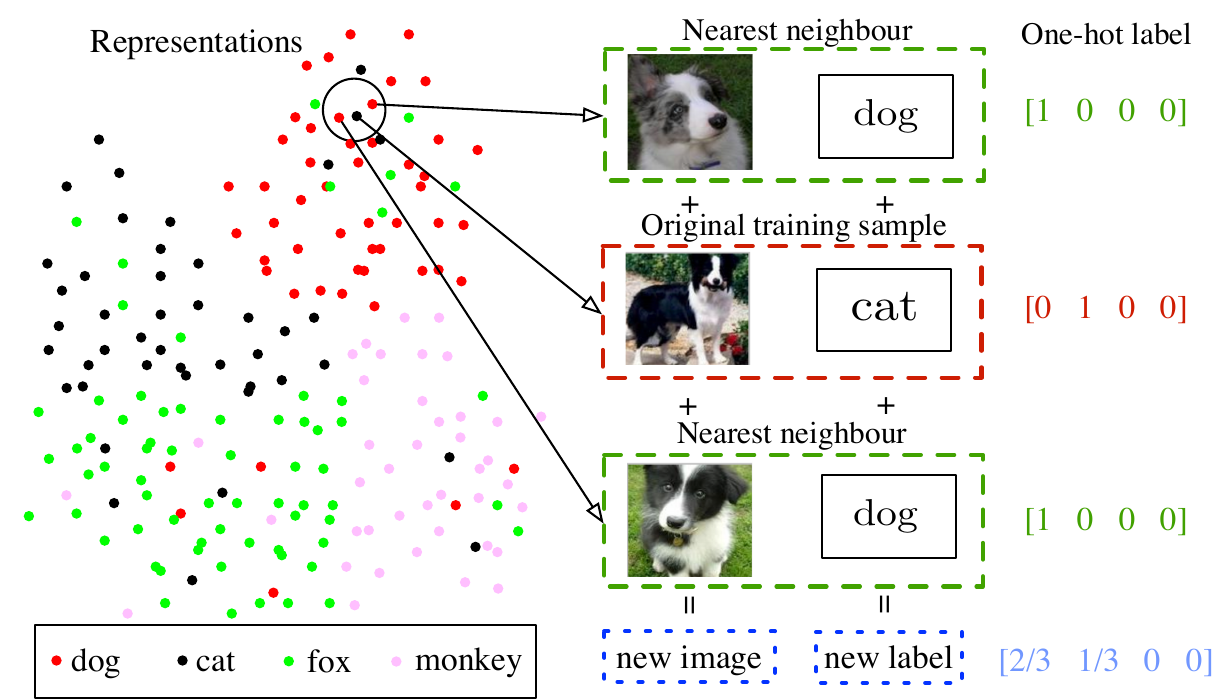}
	\end{center}
	\caption{The Synthetic samples are generated by mixing itself with its nearest neighbours based on the representations. }
	\label{fig:introduction}
\end{figure}

Given a training set consisting of clean samples and mislabeled samples, a common approach to mitigate the negative impact of noisy labels is to detect the mislabeled samples and eliminate them in the first stage, then train a new classifier with the remaining clean samples in the second stage \cite{thulasidasan2019combating,kim2019nlnl,huang2019o2u}. However, the first stage's filter mechanism for distinguishing the mislabeled samples from the others is critical to the second stage's classification performance. If the filtering mechanism only removes a few mislabeled samples, the unfiltered mislabeled samples still affect the (supervised) loss and deteriorate the classification performance. On the other hand, if too many samples, including clean samples, are eliminated, the remaining data may not be rich enough to generalize to held-out data effectively in the second stage. Therefore, is it possible to robustly train DNNs on noisy data without discording the informative training samples?

In this paper, we extend the work \cite{lu2021mixnn} for combating the negative influence of noisy labels. Our main idea is to generate new synthetic samples that effectively hide the information from the noisy labels, allowing robustly training the DNNs. Specifically, for each training sample, we search its $K$ nearest neighbours based on the learned representations. Then we linearly combine the samples with their nearest neighbours in terms of their images and labels to create synthetic samples. By training DNNs with these synthetic samples, our approach effectively reduces the noisy supervision from mislabeled samples and enhances the task performance. Take the case in Fig. \ref{fig:introduction} as an example -- Suppose we simply average the three images and their labels, then the new image is a mixed dog image and the new label is a smoothed label with the probability of $2/3$ for dog class and $1/3$ for cat class. It is clear to observe that training DNNs with the synthetic samples is more reasonable than train them with the original ones (e.g. a dog image with a wrong label cat).


Instead of simply averaging the samples with its nearest neighbours, we assign dynamic weights to these selected samples and convexly combine them to generate the synthetic samples. Ideally, the weights for clean samples should be large to preserve the correct information while small for mislabeled samples to suppress the wrong supervision. Previous work \cite{arpit2017closer} has observed that DNNs learn the clean pattern before memorizing the complex noisy pattern during training. Specifically, DNNs learn from clean samples at ease and receive inconsistent error supervision from the wrong samples before over-fitting to the entire dataset. Therefore, the predictions and given labels are likely to be consistent on clean samples and inconsistent on wrong samples, resulting in the separation of their loss values in the early learning phase. Based on this observation, we propose to estimate the weights by fitting a two-component Gaussian Mixture Model \cite{permuter2006study} to the per-sample loss distribution and calculate the posterior probability of the loss value to measure whether a sample is clean or not, allowing the mixing strategy to be dynamic in training.

To enhance the robustness of our approach, we propose a strategy to estimate target distribution based on model predictions and given noisy labels. The original noisy labels in our loss function are substituted with new soft labels, which has already been proven to effectively improve the generalization in knowledge distillation. In our scenario, the label quality is improved as the wrong labels have been gradually corrected. Subsequently, the labels of synthetic samples become more accurate, providing a more stable supervisory signal during training.

Our main contributions are summarized as follows:
\begin{itemize}
	\item We provide insights into learned representation in early learning stage. Based on it, we propose to generate new synthetic samples to robustly train DNNs, by aggregating the original samples with their top-$K$ nearest neighbours.
	\item We estimate the dynamic weights by unsupervised learning in the creation of synthetic samples. The weights are proportional to the clean probability of samples, thereby maintaining correct information while eliminating the wrong information from the mislabeled samples.
	\item We propose a strategy to gradually estimate the soft labels in the loss function, yields enhancing the performance in the presence of extreme label noise.
	\item We demonstrate that the proposed method outperforms the state-of-the-art methods on two standard benchmarks with simulated label noise and two real-world noisy datasets. We also provide extensive ablation study and empirical analyses to verify the effectiveness of different components.
\end{itemize}

The remainder of this paper is organized as follows: Section \ref{sec:related} introduces the related work on classification with noisy labels. Section \ref{sec:pre} explores the early learning phenomenon and explains the failure of using cross-entropy loss when learning with noisy labels. Section \ref{sec:method} describes each part of our method with detailed explanations. In Section \ref{sec:case_discussion}, we conduct a case study towards a better understanding of our approach and discuss the possible limitations. Section \ref{sec:exp} details the experiments including a comparison with other state-of-art methods and several analyses. Finally, we conclude the paper in Section \ref{sec:conclusion}.

\section{Related work}
\label{sec:related}
\noindent\textbf{Robust loss functions.} These studies \cite{ghosh2017robust,zhang2018generalized,wang2019symmetric,ma2020normalized} focus on developing noise-tolerant loss functions. For instance, Ghosh et al. \cite{ghosh2017robust} have proven mean absolute error loss is a noise-tolerant loss function. GCE \cite{zhang2018generalized} applies a Box-Cox transformation to probabilities which behaves like a generalized mixture of MAE and CE. 

\noindent\textbf{Loss correction and Label correction.} These approaches either iteratively relabel the noisy labels with their own predictions \cite{tanaka2018joint,wang2018iterative,yi2019probabilistic} or estimate the noise transition matrix \cite{patrini2017making}. For example, Patrini et al. \cite{patrini2017making} estimate the noise transition matrix and equally treat all samples to correct the loss. Joint-optim \cite{tanaka2018joint} iteratively updates the labels with soft or hard pseudo-labels of current predictions. 

\noindent\textbf{Sample selection by Curriculum Learning.} These methods \cite{jiang2017mentornet,malach2017decoupling,han2018co,lu2022ensemble} effectively train DNNs by selecting samples through a meaningful order (i.e. from easy samples to the hard ones). MentorNet \cite{jiang2017mentornet} pre-trains a mentor network for selecting clean samples to guide the training of the student network. Co-teaching \cite{han2018co} symmetrically train two networks by selecting small-loss instances in a mini-batch to teach the other. 

\noindent\textbf{Semi-supervised learning and meta-learning.} These methods either apply semi-supervised learning techniques after explicitly differentiating noisy samples from training data \cite{li2020dividemix,kim2019nlnl,nguyen2020self} or use meta-learning \cite{li2019learning}. For example, Li et al. \cite{li2020dividemix} divide the training data into clean and noisy ones, then train two networks with a semi-supervised algorithm MixMatch \cite{berthelot2019mixmatch}. SELF \cite{nguyen2020self} progressively filters out mislabeled samples with a semi-supervised approach.

\noindent\textbf{Regularization.} ELR \cite{liu2020early} prevents memorization of mislabeled samples by using an explicit regularizer. CAR \cite{lu2021confidence} proposes a confidence adaptive regularization to prevent memorization of noisy labels by scaling the gradient.


In contrast to the aforementioned literature, our method trains DNNs on noisy labels without: 1) consulting any clean subset; 2) eliminating training samples; 3) applying augmentation techniques from semi-supervised learning; 4) using any prior information. Specifically, we train DNNs with a new synthetic set created by aggregating the original samples with their top-$K$ nearest neighbours, which prevents DNNs from overfitting to noisy labels during training.

\section{Weakness of Cross Entropy Loss}
\label{sec:pre}
Our goal is to develop an algorithm for training DNNs classifier that performs well on the clean test set despite the presence of noisy labels in the training set. Before introducing our method, we briefly introduce the preliminary of classification with noisy labels. Then we provide some deep insights to the memorization procedure of DNNs and analyze the gradient coefficient to explain the weakness of using cross-entropy loss when learning with noisy labels.

\begin{figure}[t]
	\begin{center}
		\subfigure[Accuracy]{
			\includegraphics[width=0.46\columnwidth]{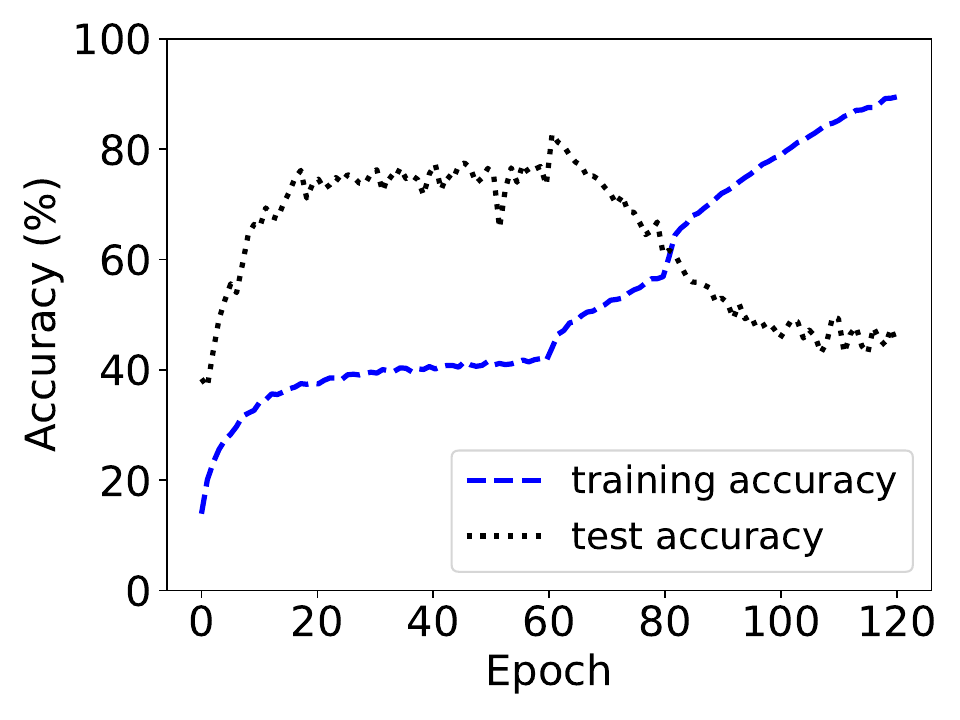}}
		\subfigure[Gradient Coefficient]{
			\includegraphics[width=0.46\columnwidth]{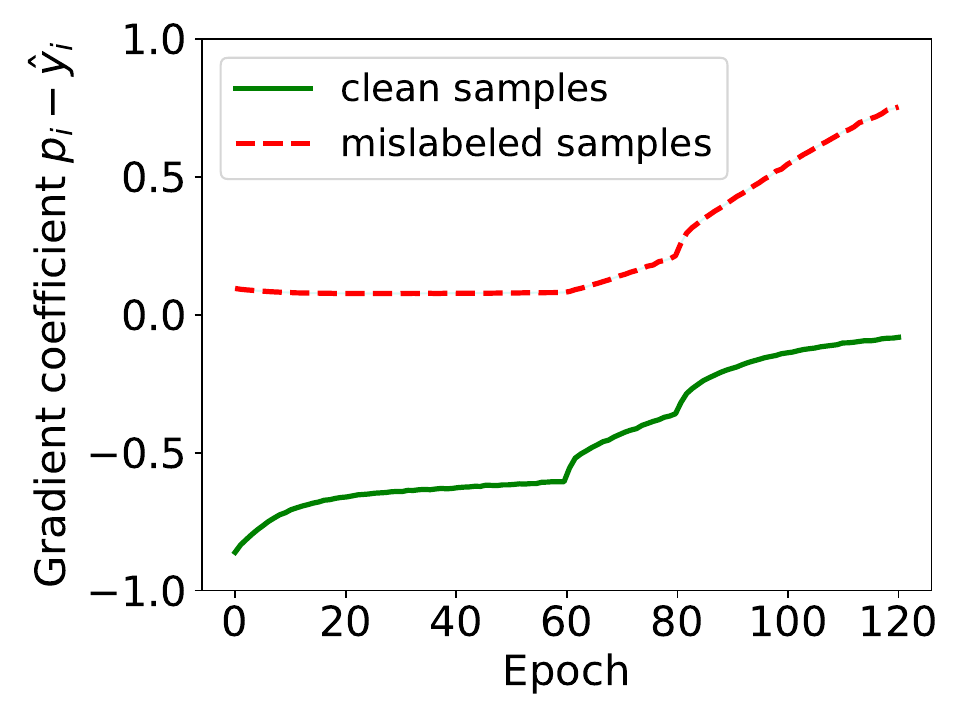}}
	\end{center}
	\caption{We train ResNet34 \cite{he2016deep} on CIFAR-10 dataset with 60\% symmetric label noise using CE loss. (a). The train and test accuracy vs. the number of training epochs. (b). The gradient coefficient $\bm{p}_{i}-\bm{\hat{y}}_{i}$ of clean and mislabeled samples vs. the number of training epochs.}
	\label{fig:memorization}
\end{figure}

\subsection{Preliminary}
Consider the $C$-class classification problem, we have a noisy training set $\hat{D}=\{(\bm{x}_{i},\bm{\hat{y}}_{i})\}^{N}_{i=1}$, where $\bm{x}_{i}$ is an input and $\bm{\hat{y}}_{i} \in \{0,1\}^{C}$ is the one-hot vector corresponding to $\bm{x}_{i}$. Note that the ground truth label $\bm{y}_{i}$ is unobservable, and the observable noisy label $\bm{\hat{y}}_{i}$ is of certain probability to be incorrect. The classification model maps each input $\bm{x}_{i}$ to a $C$-dimensional logits using a deep neural network model $\mathcal{N}_{\Theta}$ and then feeds the logits into a softmax function to produce $\bm{p}_{i}$ of the conditional probability of each class.
\begin{align}
	\label{eq:regular_p}
	\bm{p}_{i}=\mathtt{softmax}(\mathcal{N}_{\Theta}(\bm{x}_{i}))=\frac{e^{\mathcal{N}_{\Theta}(\bm{x}_{i})}}{\sum_{c=1}^{C}e^{(\mathcal{N}_{\Theta}(\bm{x}_{i}))_{c}}}.
\end{align}
$\Theta$ denotes the parameters of the DNNs and $(\mathcal{N}_{\Theta}(\bm{x}_{i}))_{c}$ denotes the $c$-th entry of logits $\mathcal{N}_{\Theta}(\bm{x}_{i})$. Traditionally, the model $\mathcal{N}_{\Theta}$ is trained via the cross-entropy (CE) loss to measure how well the model fits the training samples.
\begin{align}
	\label{eq:ce}
	\mathcal{L}_{ce}(\hat{D},\Theta)=-\frac{1}{N}\sum_{i=1}^{N}\{\ell_{ce}\}^{i}=-\frac{1}{N}\sum_{i=1}^{N}\bm{\hat{y}}_{i}^{\top}\log(\bm{p}_{i}).
\end{align}

\begin{figure*}[h]
	\begin{center}
		\includegraphics[width=0.8\linewidth]{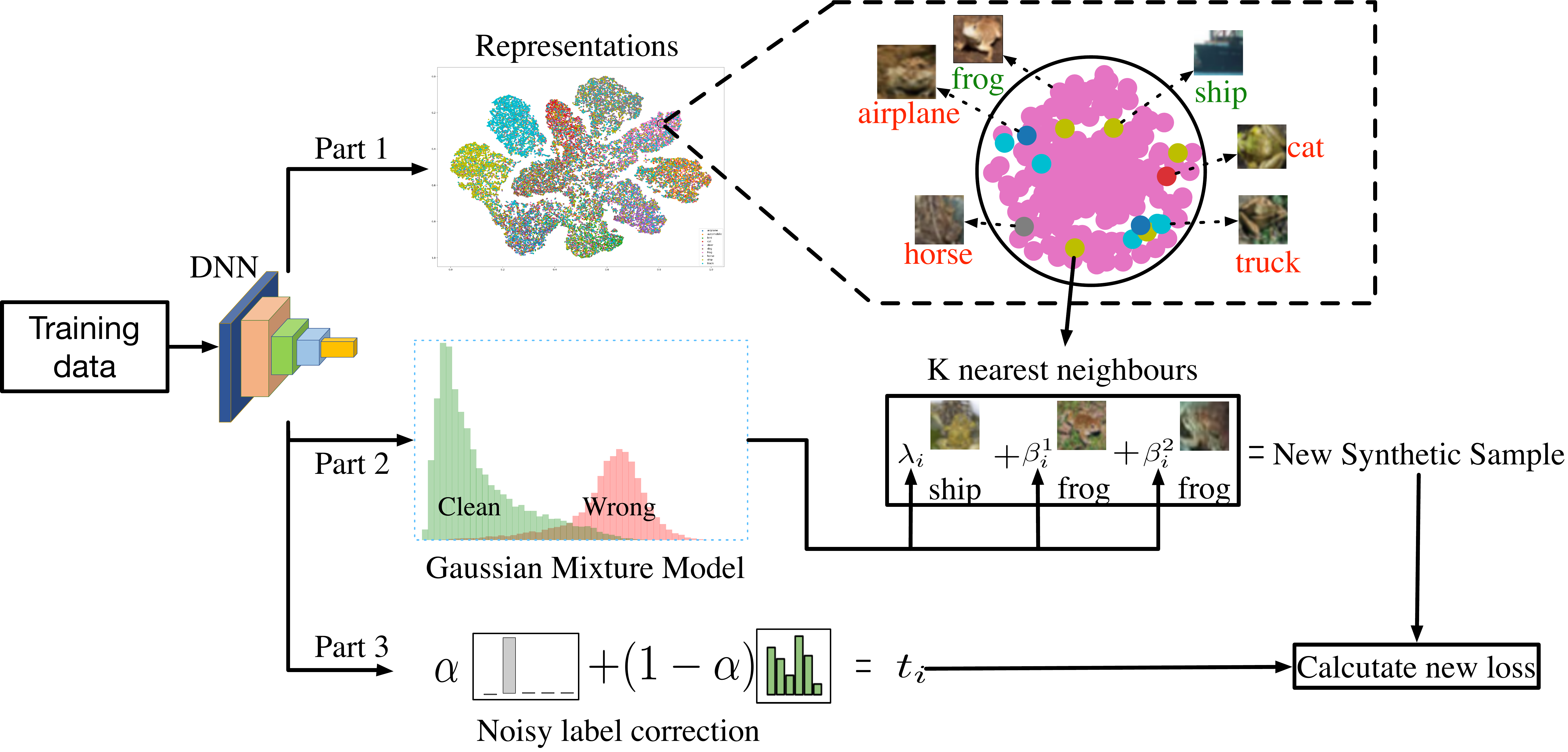}
	\end{center}
	\caption{The proposed method MixNN consists of three parts. Part 1: Based on the learned representations from the penultimate layer, we calculate each training sample's approximate $K$-nearest neighbours by using Hierarchical Navigable Small World (HNSW) graph. Part 2: We aggregate the original sample with its $K$-nearest neighbours by using the dynamic weights estimated from a Gaussian Mixture Model that learned on per-sample loss distribution. Part 3: We gradually correct the noisy labels through an exponential moving average strategy.}
	\label{fig:framework}
\end{figure*}

\subsection{Memorization Procedure of CE under Noisy Labels}
\label{sec:early}
When trained with noisy labels, the overparameterized DNNs have been observed to first fit the training data with clean labels during an \emph{early learning} stage, before eventually memorizing the examples with wrong labels \cite{arpit2017closer}. Reflected on training and test accuracy in Figure \ref{fig:memorization}, the model achieves maximum test accuracy before achieving the highest training accuracy. During training, the model starts by learning to predict the true labels for correctly labeled training samples. Thus it can predict correct labels for clean test data. However, with the increasing number of training epochs, the model begins making incorrect predictions as it memorizes the mislabeled samples. In addition to the empirical observations, recent study \cite{liu2020early} has theoretically proven such an early learning phenomenon also occurs in a simple linear model.

\subsection{Gradient Analysis of CE}
\label{sec:grad_inversion}
To further explain why the model memorizes noisy labels eventually, we drive the gradient of cross entropy loss with respect to $\Theta$ as follows:
\begin{align}
	\label{eq:grad}
	\nabla \mathcal{L}_{ce}(\hat{D},\Theta)=-\frac{1}{N}\sum_{i=1}^{N}\nabla\mathcal{N}_{\Theta}(\bm{x}_{i})(\bm{p}_{i}-\bm{\hat{y}}_{i}),
\end{align}
where $\nabla\mathcal{N}_{\Theta}(\bm{x}_{i})$ is the Jacobian matrix of the neural network logits for the $i$-th input with respect to $\Theta$. Assume in clean training data scenario, $\bm{p}_{i}-\bm{y}_{i}$ of true class entry will always be negative and the rest entries are positive. Therefore, performing stochastic gradient descent increases the probability of true class and reduces the residual probabilities at other classes, which ensures the learning to continue on true class. However, in the noisy label scenario, if $c$ is the true class, but $c$-th entry of noisy label $(\hat{\bm{y}}_{i})_{c}=0$, then the contribution of the $i$-th sample to $\nabla \mathcal{L}_{ce}(\Theta)$ is reversed (i.e. $(\bm{p}_{i}-\hat{\bm{y}}_{i})_{c}$ should be negative but get positive instead). In the meanwhile, the entry corresponding to the impostor class $c'$, is also reversed because $(\hat{\bm{y}}_{i})_{c'}=1$. Thus, for clean samples, the cross-entropy term $\bm{p}_{i}-\bm{\hat{y}}_{i}$ tends to vanish (i.e. closer to zero) after the early learning stage because $\bm{p}_{i}$ is close to $\bm{y}_{i}$. For mislabeled samples, the cross-entropy term $(\bm{p}_{i}-\hat{\bm{y}}_{i})_{c}$ is positive, allowing them to dominate the gradient. The right plot in Figure \ref{fig:memorization} shows the change of gradient coefficient $(\bm{p}_{i}-\hat{\bm{y}}_{i})_{c}$ in training. The gradients of clean samples dominate at the beginning, and then are gradually suppressed by the gradients of mislabeled samples. Therefore, performing stochastic gradient descent eventually results in the memorization of whole training data including mislabeled samples. 

\section{Methodology}
\label{sec:method}
We name our framework \textbf{MixNN} as it \textbf{Mix}es each training sample with its \textbf{N}earest \textbf{N}eighbours to train the DNNs. A diagram of our framework is shown in Fig. \ref{fig:framework}. Generally, MixNN consists of three parts (boxed with different colors). In this section, we first explore the representation distributions in the early learning stage. Then we describe the details in MixNN, including mixing with nearest neighbours, $K$-approximate nearest neighbour search, weight estimation, and noisy labels correction. 

\subsection{Representation Distributions}
\label{sec:rep_d}
In the noisy label scenario, the training data consist of clean samples and mislabeled samples. The goal of MixNN is to prevent the model from memorizing mislabeled samples while continually learning from clean samples. To study whether the learned representations are corrupted due to label noise, we plot the t-SNE graph \cite{van2008visualizing} of learned representations (i.e. the embeddings from penultimate layer) in the early learning stage in Fig. \ref{fig:framework} Part 1. As we can see, the learned representations of majority clean samples still congregate in their true classes, while the representations of mislabeled samples disperse in all classes. For clear illustration, we zoom in a random region in frog (pink) class and display the samples in the right as an example. In this region, we observe that most of the frog images have learned correct representations, only a few samples from other classes have learned `corrupted' representations (e.g. the ship image shouldn't have a fog representation). Besides, most of the mislabeled samples in frog class are also frog images though attached wrong labels from other classes. In our experiments, we find that 88.96\% and 82.8\% mislabeled samples' representations still congregate in their true class for CIFAR-10 with 40\% and 60\% label noise respectively. 

This observation motivates us to consider whether we can refer the correct information from the mislabeled samples' nearest neighbours. To achieve our goal, we generate a new training set where each sample is mixed with its $K$ nearest neighours. Ideally, the correct image features are most likely to be preserved while the negative influence of noisy labels can be mitigated, yields achieving robustness to label noise. 

\subsection{Mixing with Nearest Neighbours}
Our main idea is to use the correct knowledge from nearest neighbours to mitigate the detrimental impact of noisy labels. For each training sample $\bm{x}_{i}$ in a mini-batch, we generate a synthetic training sample $\bm{\tilde{x}}_{i}$ by convex linearly combining it with its $K$ nearest neighbours. We denote $K$ nearest neighbours of a training sample $(\bm{x}_{i},\bm{\hat{y}}_{i})$ as $\Phi^{K}(\bm{x}_{i})=\{(\bar{\bm{x}}_{i}^{k},\bar{\bm{y}}_{i}^{k})\}_{k=1}^{K}=\{(\bar{\bm{x}}_{i}^{1},\bar{\bm{y}}_{i}^{1}),(\bar{\bm{x}}_{i}^{2},\bar{\bm{y}}_{i}^{2}),\dots,(\bar{\bm{x}}_{i}^{K},\bar{\bm{y}}_{i}^{K})\}$. Hence, we define the image of a synthetic training sample as
\begin{align}
	\label{eq:mix_x}
	\bm{\tilde{x}}_{i}= \lambda_{i} \bm{x}_{i} + \sum_{k=1}^{K}\beta^{k}_{i}\bar{\bm{x}}^{k}_{i}, \quad \sum_{k=1}^{K}\beta_{i}^{k}=1-\lambda_{i},
\end{align}
where $\lambda_{i}$ is a dynamic scalar value denoting the weight of original training sample. $\beta^{k}_{i}$ denotes the weight of $k$-th nearest neighbour. We have $\lambda_{i} +\sum_{k=1}^{K}\beta_{i}^{k}=1$ to ensure the synthetic training sample still follows the same distribution of original sample after normalization. Similarly, we calculate the new label $\bm{\tilde{y}}_{i}$ of the new sample by 
\begin{align}
	\label{eq:mix_y}
	\bm{\tilde{y}}_{i}= \lambda_{i} \bm{\hat{y}}_{i} + \sum_{k=1}^{K}\beta_{i}^{k}\bar{\bm{y}}^{k}_{i}, \quad \sum_{k=1}^{K}\beta_{i}^{k}=1-\lambda_{i}.
\end{align}
We then train our model with the synthetic training set $\tilde{D}=\{(\bm{\tilde{x}}_{i},\bm{\tilde{y}}_{i})\}^{N}_{i=1}$.  Based on $\tilde{D}$, we use the cross-entropy loss as the measure of how well the model fits the $\tilde{D}$. We denote 
\begin{align}
	\label{eq:tildep}
	\bm{\tilde{p}}_{i}=\mathtt{softmax}(\mathcal{N}_{\Theta}(\bm{\tilde{x}}_{i}))=\frac{e^{\mathcal{N}_{\Theta}(\bm{\tilde{x}}_{i})}}{\sum_{c=1}^{C}e^{(\mathcal{N}_{\Theta}(\bm{\tilde{x}}_{i}))_{c}}}.
\end{align}
Thus, the new loss $\mathcal{L}(\tilde{D},\Theta)$ becomes
\begin{align}
	\label{eq:total_loss}
	&-\frac{1}{N}\sum_{i=1}^{N}\bm{\tilde{y}}^{\top}_{i}\log(\bm{\tilde{p}}_{i})\nonumber\\
	=&-\frac{1}{N}\sum_{i=1}^{N}(\lambda_{i} \bm{\hat{y}}_{i} + \sum_{k=1}^{K}\beta_{i}^{k}\bar{\bm{y}}^{k}_{i})^{T}\log(\bm{\tilde{p}}_{i}) \nonumber \\
	=&\underbrace{-\frac{1}{N}\sum_{i=1}^{N}\lambda_{i}\bm{\hat{y}}^{\top}_{i}\log(\bm{\tilde{p}}_{i})}_\text{first term}\underbrace{-\frac{1}{N}\sum_{i=1}^{N}(\sum_{k=1}^{K}\beta_{i}^{k}\bar{\bm{y}}^{k}_{i})^{\top}\log(\bm{\tilde{p}}_{i}).}_\text{second term}
\end{align}
Compared to CE loss in Eq. (\ref{eq:ce}), the new loss is more resistant to noisy labels. In Eq. (\ref{eq:total_loss}), the first term is similar to the CE loss, except it is weighted by $\lambda_{i}$. The target in the second term is mixed by labels from $K$ nearest neighbours. In Section \ref{sec:weight_e}, we will generate $\lambda_{i}\rightarrow 1$ for a clean sample $\bm{x}_{i}$, causing $\sum_{k=1}^{K}\beta_{i}^{k}\rightarrow 0$. Therefore, the new loss reduces to exact CE loss for clean samples. For mislabeled samples, since $\lambda_{i}\rightarrow 0$, the new loss relies on the second term. According to Section \ref{sec:rep_d}, labels of $K$ nearest neighbours $\bar{\bm{y}}^{k}_{i}$ are likely to be the correct labels. The second term penalizes model prediction that are inconsistent with mixed target $\sum_{k=1}^{K}\beta_{i}^{k}\bar{\bm{y}}^{k}_{i}$, mitigating the impact of noisy labels on training DNNs.

\subsection{Approximate Nearest Neighbour Search}
In our method, we search the $K$-Nearest Neighbors (KNN) for each training sample based on the learned representation. Assume the learned representation of a training sample is a query vector. A naive approach to performing exact KNN search is to directly compute the distances (e.g. Euclidean distance and Cosine distance) between the query and every element in the training set. Hence, the complexity of the naive approach is $O(dN)$, where $N$ is the size of training set and $d$ is the dimension of representation vector. 

Previous study \cite{har2012approximate} has demonstrated that exact KNN search solutions may offer a substantial search speedup only in the case of relatively low dimentional data (e.g. $d<20$) due to ``curse of dimensionality''. For instance, the complexity of KNN search in KD-tree \cite{bentley1975multidimensional} is $O(2^{d}\log(N))$ which is exponential to the dimension of representations. In our case $d$ can be large. For example, the dimension of representation in penultimate layer for ResNet34 \cite{he2016deep} is 512. Therefore, it is inefficient to directly calcuate the exact KNN.

To overcome this problem, a concept of Approximate Nearest Neighbours Search (ANNS) \cite{har2012approximate} was proposed, which relaxes the condition of the exact search by allowing a small number of errors. The quality of an inexact search is defined as the ratio between the number of found true nearest neighbours and $K$. 
In this paper, we adopt the Hierarchical Navigable Small World (HNSW) graph \cite{malkov2018efficient} as our search index. It is a fully graph based incremental ANNS structure that can offer a superior logarithmic complexity scaling. The search index in HNSW is a multi-layered structure where each layer is a proximity graph. Each node in the graph corresponds to one of the representations. A nearest neighbour search in HNSW adopts a ``zooming-in'' style. It starts at an entry point node in the uppermost layer and recursively performs a greedy graph traversal in each layer until it reaches a local minimum in the bottommost one. The maximum number of connections per element in all layers can be made a constant, thus allowing a logarithmic complexity scaling of routing in a navigable small world graph. In this paper, we use Euclidean distance as the measure of similarity and the overall search complexity scaling is $O(\log(N))$. When $K=4$, the original exact KNN search costs 5.25 millisecond per sample in our experiment. When using HWSW, the search time is reduced to 0.034 millisecond per sample, which is quite efficient.

\subsection{Weight Estimation}
\label{sec:weight_e}
In mixing functions Eq. (\ref{eq:mix_x}) and Eq. (\ref{eq:mix_y}), the weight $\lambda_{i}$ indicates how confidently we can trust the original sample, whereas $\beta_{i}^{1},...,\beta_{i}^{K}$ indicates how much knowledge is referred from these nearest neighbours. Ideally, we want to preserve correct information from the clean samples while dampening the wrong information from the mislabeled samples. In other words, the weights should be able to indicate the `probability' of a training sample being correctly labeled or not.

\begin{figure*}
	\centering
	\subfigure[40\% symmetric noise]{
		\includegraphics[width=0.48\columnwidth]{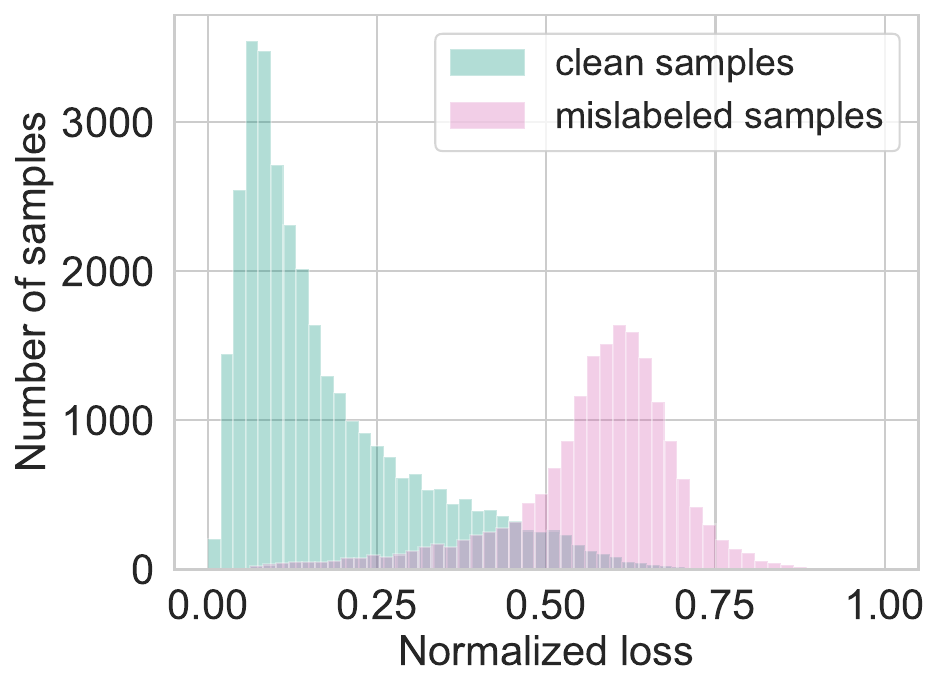}}
	\subfigure[40\% symmetric noise]{
		\includegraphics[width=0.48\columnwidth]{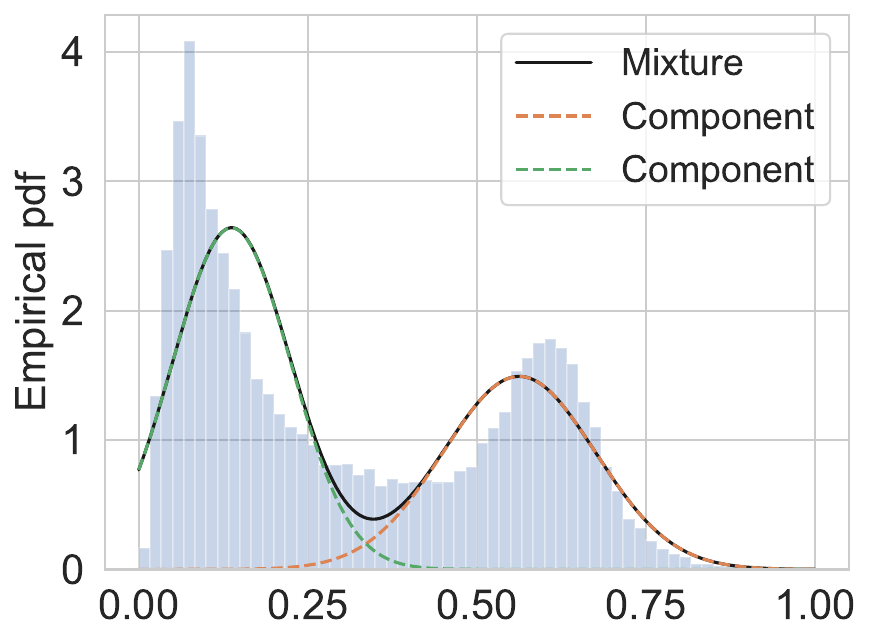}}
	\subfigure[80\% symmetric noise]{
		\includegraphics[width=0.48\columnwidth]{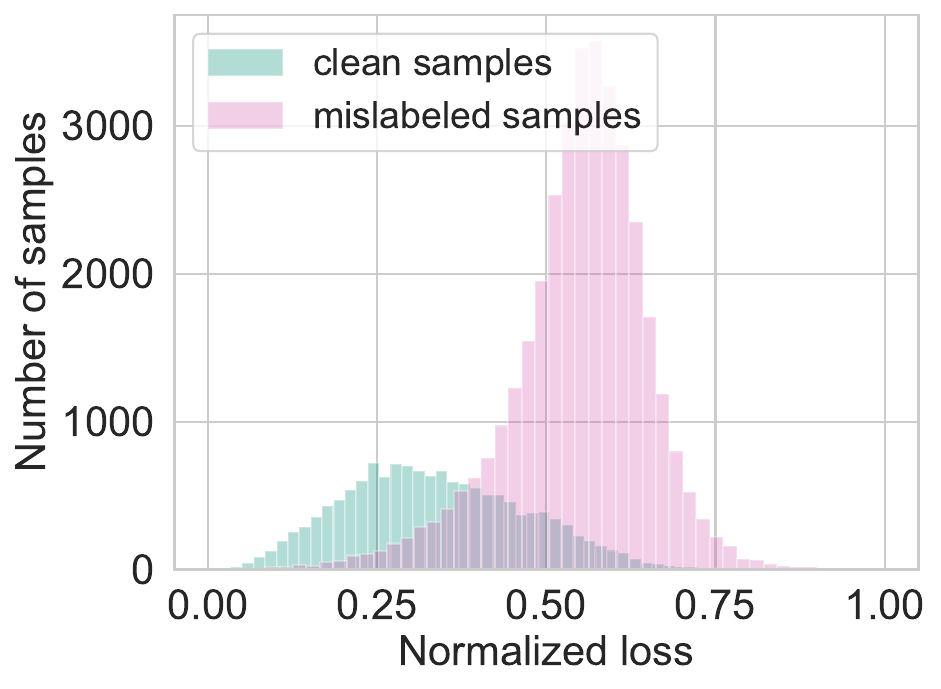}}
	\subfigure[80\% symmetric noise]{
		\includegraphics[width=0.48\columnwidth]{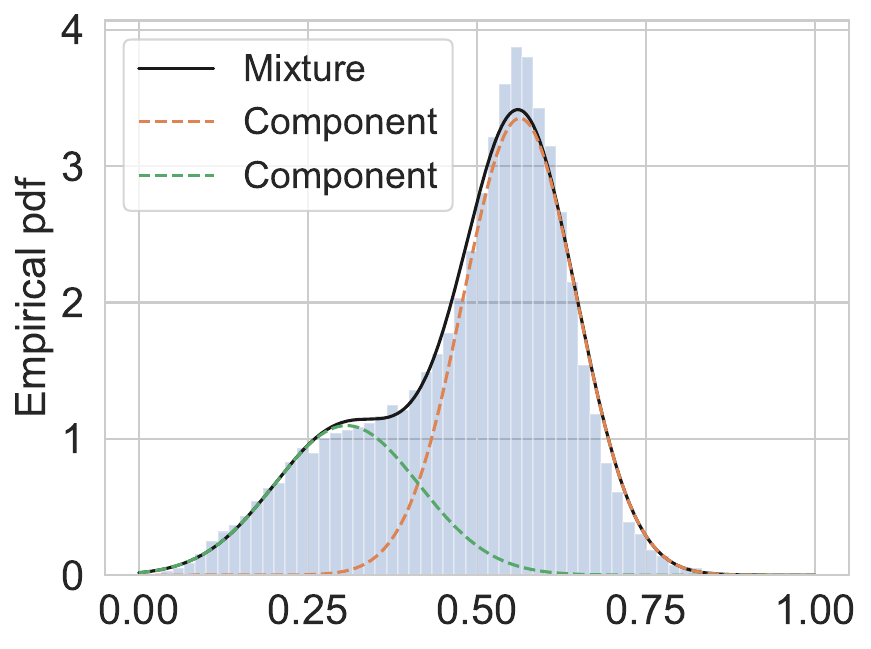}}
	\caption{Train on CIFAR-10 with 40\% and 80\% label noise after 10 epochs with cross-entropy loss. Plots (a) and (c): The ground truth normalized loss distribution. Plots (b) and (d): The pdf of mixture model and two components after fitting a two-component GMM to per-sample loss distribution.}
	\label{fig:loss_dist}
\end{figure*}

Due to the early learning phenomenon, the samples with small-loss values are more likely to be correctly labeled. Therefore, existing sample selection methods \cite{han2018co} select the clean samples according to the magnitude of loss values. In this paper, we investigate the per-sample loss distribution and find that there is a separation between the loss distribution of correctly labeled samples and loss distribution of mislabeled samples. As shown in Fig. \ref{fig:loss_dist} (a) and (c), the normalized loss values of the clean samples are in expectation smaller than the mislabeled ones. Even in the case of extreme label noise (e.g. 80\% labels are incorrect), their loss distributions can be differentiated. To estimate the probability of a sample being clean, we introduce a two-component Gaussian Mixture Model (GMM) \cite{permuter2006study} to fit the per-sample loss distribution as shown in Fig. \ref{fig:loss_dist} (b) and (d). The probability density function (pdf) of GMM with $M$ components on the per sample loss value $\ell$ can be defined as  
\begin{align}
	P(\ell)=\sum_{m=1}^{M}\pi_{m}\mathcal{G}(\ell \mid \mu_{m}, \sigma^{2}_{m}), \quad \sum_{m=1}^{M}\pi_{m}=1,
\end{align}
where $\pi_{m}$ is the coefficient for the linear convex combination of each individual pdf $\mathcal{G}(\ell \mid \mu_{m}, \sigma^{2}_{m})$. In our case, we use an Expectation-Maximization (EM) algorithm to estimate the $\pi_{m}$, $\mu_{m}$ and $\sigma_{m}^{2}$. Therefore, we obtain the probability of a sample being clean or mislabeled through the posterior probability:
\begin{align}
	P(m\mid \ell)= \frac{P(m)P(\ell\mid m)}{P(\ell)}=\frac{\pi_{m}\mathcal{G}(\ell\mid \mu_{m},\sigma^{2}_{m})}{\sum_{m=1}^{M}\pi_{m}\mathcal{G}(\ell \mid \mu_{m}, \sigma^{2}_{m})}
\end{align}
where $m=0 (1)$ indicate correct (wrong) labels. Note that we always calculate the cross-entropy loss to estimate the clean probability for all samples after every epoch. But we use our loss defined in Eq. (\ref{eq:total_loss}) for training the model which contains multiple loss terms to deal with label noise. 

While mislabeled samples benefit from combining with clean ones, clean samples are contaminated by mislabeled ones, whose training objective is incorrectly modified. The goal of mixing strategy in Eq. (\ref{eq:mix_x}) and Eq. (\ref{eq:mix_y}) is to use the dynamic weights to reduce the contribution of mislabeled samples when they are combined with correctly labeled ones. We denote the per sample loss value of $\bm{x}_{i}$ as $\ell(\bm{x}_{i})$. Thus the dynamic weights are calculated by
\begin{align}
	\label{eq:lambda}
	\lambda_{i}=\frac{P(m=0\mid \ell(\bm{x}_{i}))}{P(m=0\mid \ell(\bm{x}_{i}))+\sum_{k=1}^{K}P(m=0\mid\ell(\bm{\bar{x}}_{i}^{k}))},
\end{align}
\begin{align}
	\label{eq:beta}
	\beta_{i}^{k}=\frac{P(m=0\mid \ell(\bm{\bar{x}}_{i}^{k}))}{P(m=0\mid \ell(\bm{x}_{i}))+\sum_{k=1}^{K}P(m=0\mid\ell(\bm{\bar{x}}_{i}^{k}))}.
\end{align}
We then use the above weights to guide the generation of synthetic training sample $(\bm{\tilde{x}}_{i},\bm{\tilde{y}}_{i})$. Consider $K=1$, there are totally four mixing cases: clean-clean, clean-wrong, wrong-clean, and wrong-wrong. By using dynamic weights, it largely avoids generating the confusing input to the network in clean-wrong and wrong-clean cases, while retaining the strengths for clean-clean and wrong-wrong combinations. More discussion on these four cases is in Section \ref{sec:case_discussion}.

\subsection{Noisy Labels Correction}
\label{sec:label_correct}
Despite that the synthetic training samples set $\tilde{D}=\{(\bm{\tilde{x}}_{i},\tilde{\bm{y}}_{i})\}^{N}_{i=1}$ is better than directly using given noisy training dataset $\hat{D}=\{(\bm{x}_{i},\bm{\hat{y}}_{i})\}^{N}_{i=1}$. Nevertheless, using the noisy labels $\bm{\hat{y}}_{i}$ in Eq. (\ref{eq:mix_y}) and Eq. (\ref{eq:total_loss}) may be less effective as $\bm{\hat{y}}_{i}$ is likely to be incorrect, especially when the noise rate is extremely high. Therefore, a better estimation of ground truth label $\bm{y}_{i}$ can further improve the performance. In Fig. \ref{fig:memorization}, we observe that most predictions in early learning stage are correct. Based on this observation, we use an exponential moving average strategy to gradually estimate the soft target $\bm{t}_{i}$ by using the noisy label $\hat{\bm{y}}_{i}$ and model prediction $\bm{p}_{i}$. We update $\bm{t}_{i}$ in each epoch $E$ by
\begin{align}
	\bm{t}_{i}=\left\{ \begin{array}{ll}
		\bm{\hat{y}}_{i} & \textrm{if } E < E_{s}\\
		\alpha \bm{t}_{i} + (1-\alpha) \bm{p}_{i} & \textrm{if $E \ge E_{s}$}  \\
	\end{array} \right. 
\end{align}
where $E_{s}$ is the epoch that starts performing label correction and $0\le \alpha <1$ is the momentum. We then replace the noisy label $\hat{\bm{y}}_{i}$ in Eq. (\ref{eq:mix_y}) and Eq. (\ref{eq:total_loss}) with the estimated soft target $\bm{t}_{i}$. Consequently, using a better $\bm{t}_{i}$ facilitate the model to memorize more correctly labeled samples and to generate a better new mixed label $\tilde{\bm{y}}$. The correction accuracy will be discussed in Section \ref{sec:correction_accuracy}. Overall, put all parts together, our algorithm is described in Algorithm \ref{alg:1}.

\begin{algorithm2e} [t]
	\small
	\nonumber
	\caption{MixNN pseudocode.}
	\label{alg:1}
	\KwIn{Networks $\mathcal{N}_{\Theta}$ with parameters $\Theta$, training set $\hat{D}$, number of nearest neighbors $K$, batch size $B$, learning rate $\eta$, number of total training epochs $E_\text{max}$, $E_{s}=60$ momentum $\alpha=0.9$;}
	$\Theta= \text{ Warmup}(\hat{D}, \Theta,\mathcal{L}_{ce})$; \tcp{warmup with CE loss.} 
	\textbf{Initialize} the target $\bm{t}_{i}=\bm{\hat{y}}_{i}$ for all samples in $\hat{D}$\;
	\For{$e=1,2,\dots,E_\text{max}$}
	{
		$P(m=0\mid \ell_{ce}(\bm{x}_{i}))=\text{ GMM }(\hat{D},\ell_{ce},\Theta)$; \tcp{fit GMM to loss distibution.}
		\textbf{Obtain} $K$ Approximate Nearest Neighbors  $\Phi^{K}(\bm{x}_{i})=\{(\bar{\bm{x}}_{i}^{k},\bar{\bm{y}}_{i}^{k})\}_{k=1}^{K}$ for each $\bm{x}_{i}$ by using HNSW\;
		\textbf{Shuffle} $\hat{D}$ into $\frac{|\hat{D}|}{B}$ mini-batches \;
		\For{$n=1,2,\dots,\frac{|\hat{D}|}{B}$}
		{
			\textbf{Fetch} $n$-th mini-batch $\hat{D}_{n}$ from $\hat{D}$ \;
			\textbf{Obtain} $\bm{p}_{i}$ for each sample in $\hat{D}_{n}$ by Eq. (\ref{eq:regular_p})\;
			\If{$e\ge E_{s}$}{\textbf{Update} $\bm{t}_{i}=\alpha \bm{t}_{i} + (1-\alpha) \bm{p}_{i}\text{ for all sample in $\hat{D}_{n};$}$}
			\textbf{Estimate} the weights $\lambda_{i},\beta_{i}^{1},...,\beta_{i}^{K}$ for each sample in $\hat{D}_{n}$ by Eq. (\ref{eq:lambda}) and Eq. (\ref{eq:beta}).\;
			
			\textbf{Generate} $\tilde{D}_{n}=\{(\tilde{\bm{x}}_{i},\tilde{\bm{y}}_{i})\}_{i=1}^{B}$ by Eq. (\ref{eq:mix_x}), Eq. (\ref{eq:mix_y}) and $\Phi^{K}(\bm{x}_{i})$\;

			\textbf{Obtain} $\bm{\tilde{p}}_{i}$ for each sample in $\tilde{D}_{n}$ by Eq. (\ref{eq:tildep})\;
			
			\textbf{Calculate} the loss $\mathcal{L}(\tilde{D}_{n},\Theta)=-\frac{1}{B}\sum_{i=1}^{B}\lambda_{i}\bm{t}_{i}^{T}\log(\bm{\tilde{p}}_{i})-\frac{1}{B}\sum_{i=1}^{B}(\sum_{k=1}^{K}\beta_{i}^{k}\bar{\bm{y}}^{k}_{i})^{T}\log(\bm{\tilde{p}}_{i})$\;
			
			\textbf{Update} $\Theta=\Theta-\eta\nabla \mathcal{L}(\tilde{D}_{n},\Theta)$ \; 
			
		}
	}
	\textbf{Output} $\Theta$.
\end{algorithm2e}

\section{Case Discussion}
\label{sec:case_discussion}
\begin{figure}[t]
	\begin{center}
		\includegraphics[width=0.7\linewidth]{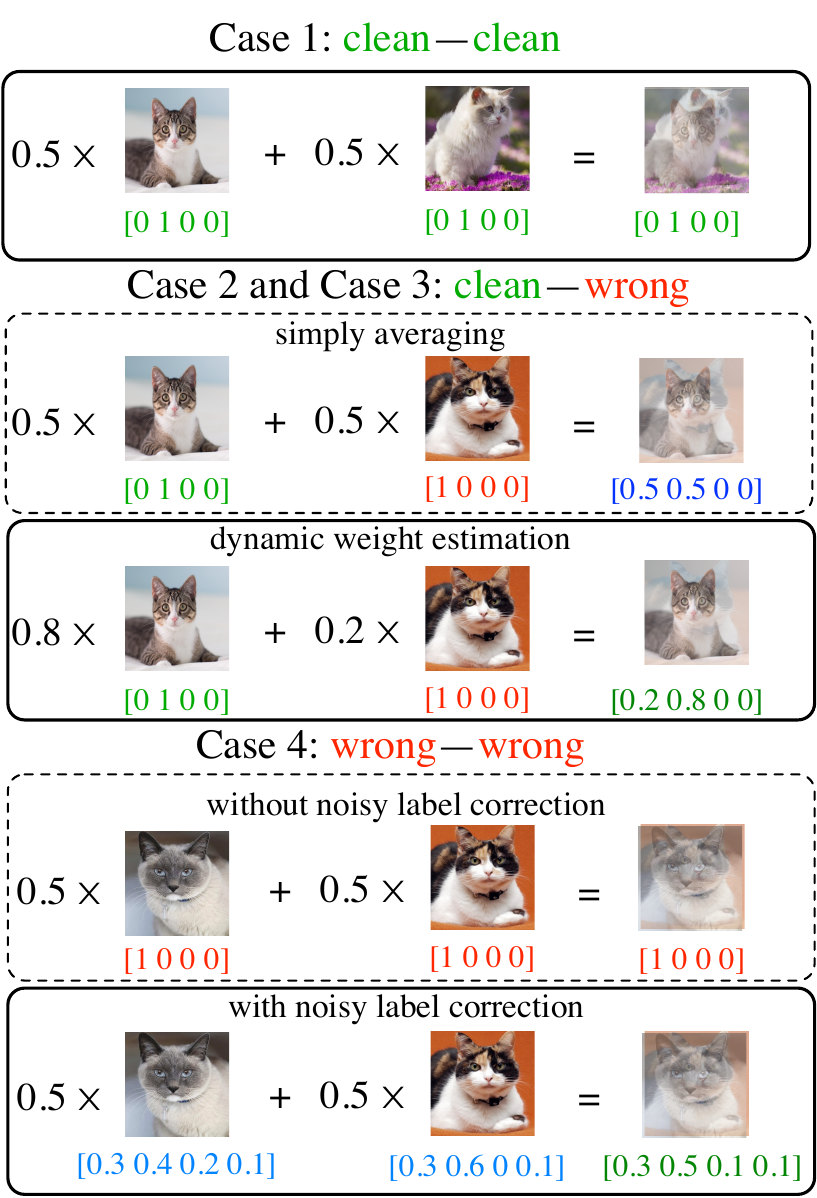}
	\end{center}
	\caption{A toy example for illustrating the different cases in MixNN. We use the one-hot label vector and each entry indicates four different classes (i.e. dog, cat, fox and monkey).}
	\label{fig:cases}
\end{figure}

To explain how MixNN works, we discuss the possible cases when mixing with one nearest neighbour ($K=1$). We also provide a toy example in Fig. \ref{fig:cases} to better illustrate each case. The synthetic samples $(\bm{\tilde{x}}_{i},\bm{\tilde{y}}_{i})$ are generated by the following four ways.

\noindent \textbf{Case 1}: \emph{Both $(\bm{x}_{i},\hat{\bm{y}}_{i})$ and its nearest neighbour $(\bar{\bm{x}}_{i}^{1},\bar{\bm{y}_{i}}^{1})$ are clean samples.} Based on the analysis of representation distributions in the early learning phase, both $\bm{x}_{i}$ and $\bar{\bm{x}}_{i}^{1}$ are most likely to be the images from the same class. In this case, the new mixed input is a convex linear combination of two similar images, i.e., $\bm{\tilde{x}}_{i}=\lambda_{i}\bm{x}_{i}+\beta^{1}_{i}\bar{\bm{x}}_{i}^{1}$. And its corresponding label is $\bm{\tilde{y}}_{i}=\lambda_{i}\bm{\hat{y}}_{i}+\beta^{1}_{i}\bar{\bm{y}}^{1}_{i}$. Since $\bm{\hat{y}}_{i}=\bar{\bm{y}}^{1}_{i}$ and $\lambda_{i}+\beta^{1}_{i}=1$, then the mixed label $\bm{\tilde{y}}_{i}=\bm{\hat{y}}_{i}=\bar{\bm{y}}^{1}_{i}$. Therefore, the new sample $(\bm{\tilde{x}}_{i},\bm{\tilde{y}}_{i})$ is similar to the sample after applying augmentation strategy \cite{zhang2017mixup} which encourages the model to behave linearly in-between training samples, resulting in reducing the number of undesirable oscillations when predicting hard samples. The DNNs trained with such samples will be better calibrated \cite{thulasidasan2019mixup}. In other words, the prediction softmax scores are much better indicators of the actual likelihood of predictions, which avoids producing the overconfident wrong predictions and improves the estimation of dynamic weights in Section \ref{sec:weight_e}.

\noindent \textbf{Case 2}: \emph{$(\bm{x}_{i},\hat{\bm{y}}_{i})$ is clean sample and its nearest neighbour $(\bar{\bm{x}}_{i}^{1},\bar{\bm{y}_{i}}^{1})$ is mislabeled sample.} Similarly, the new mixed input is most likely to be a combination of two images from the same class, while their labels are inconsistent. In Fig. \ref{fig:cases} case 2, the mixed label becomes an ambiguous target if simply averaging the samples. However, with estimated dynamic weights, the mixed label is determined and correct. Our method to generate the mixed label is similar to label smoothing \cite{szegedy2016rethinking} which scales and translates the original noisy label $\hat{\bm{y}}_{i}$ to $(1-\gamma)\hat{\bm{y}}_{i}+\gamma/C$, but preserves the label with maximal probability when $\gamma<1$. Different from the label smoothing that uses the fixed uniform distribution to scale the noisy labels, our approach adopts the dynamic weights, where $\lambda,\beta^{1},\dots,\beta^{K}$ are learned from data, to adaptively adjust the mixed label for better performance.

\noindent \textbf{Case 3}: \emph{$(\bm{x}_{i},\hat{\bm{y}}_{i})$ is mislabeled sample and its nearest neighbour $(\bar{\bm{x}}_{i}^{1},\bar{\bm{y}_{i}}^{1})$ is clean sample.} This case is similar to case 2, so we do not further discuss it.

\noindent \textbf{Case 4}: \emph{Both $(\bm{x}_{i},\hat{\bm{y}}_{i})$ and its nearest neighbour $(\bar{\bm{x}}_{i}^{1},\bar{\bm{y}_{i}}^{1})$ are mislabeled samples.} In this case, the mixed input is combined with two images from the same class, but their labels are both incorrect. Thus the mixed sample is not promised to improve the generalization capacity as shown in Fig. \ref{fig:cases}. However, the noisy labels are very likely to be corrected after using label correction (more discussions in Section \ref{sec:correction_accuracy})). Therefore, the mixed label can be a reasonable target compared to the original noisy one. 


\section{Experiments}
\label{sec:exp}
In this section, we first test the efficacy of the proposed method on two benchmark datasets with simulated label noise. Then we evaluate the performance of MixNN on the real-world datasets which contain more sophisticated label noise. We also provide ablation study and qualitative analyses to investigate the effect of different components. All experiments are implemented in Pytorch and run on a single NVIDIA A100 GPU. 


\subsection{Effectiveness on Simulated Label Noise}
We conduct the experiments with simulated label noise on the two benchmarks \emph{CIFAR-10} and \emph{CIFAR-100} \cite{krizhevsky2012imat}.

\begin{figure}[t]
	\begin{center}
		\includegraphics[width=0.7\linewidth]{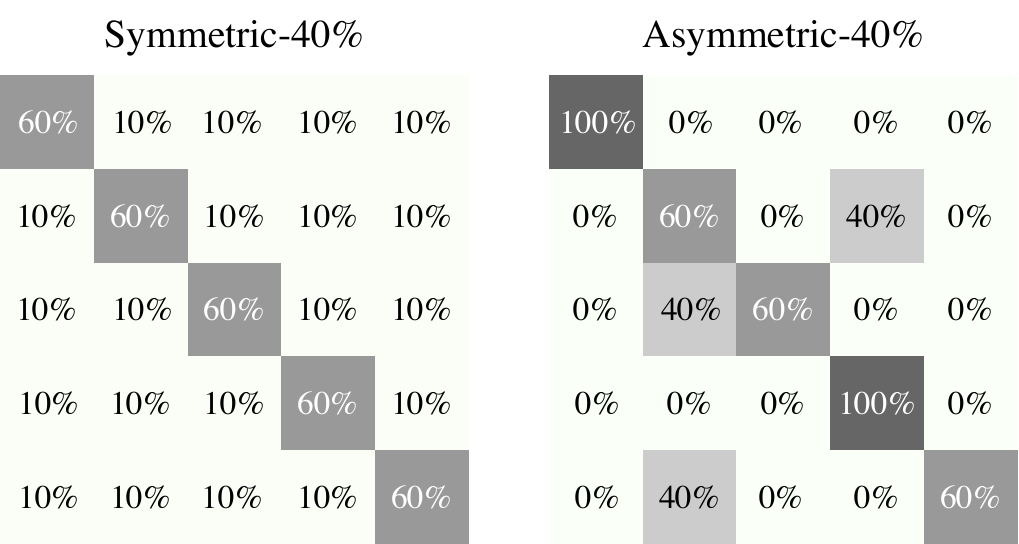}
	\end{center}
	\caption{Example of label transition matrix $\mathbf{Q}$ (taking 5 classes and noise ratio $\varepsilon=0.4$ as an example).}
	\label{fig:noise_transition}
\end{figure}

\noindent\textbf{Label Noise Simulation.} Given CIFAR-10 and CIFAR-100 are initially clean, we follow the way in \cite{patrini2017making} to corrupt these two benchmarks manually by label transition matrix $\mathbf{Q}$, where $\mathbf{Q}_{ij}=\Pr[\hat{y}=j \mid y=i]$ denotes the probability that noisy label $\hat{y}$ is flipped from clean label $y$. Generally, the matrix $\mathbf{Q}$ has two representative label noise models. (1) Symmetric noise is generated by uniformly flipping labels in each class to one of the other class labels with probability $\varepsilon$. (2) Asymmetric noise is a simulation of fine-grained classification with noisy labels in the real world, where the annotators are more likely to make mistakes only within very similar classes. Fig. \ref{fig:noise_transition} shows an example of label transition matrix $\mathbf{Q}$ for above two label noise models. In this paper, the asymmetric noisy labels are generated by flipping \emph{truck} $\rightarrow$ \emph{automobile}, \emph{bird} $\rightarrow$ \emph{airplane}, \emph{deer} $\rightarrow$ \emph{horse} and \emph{cat} $\leftrightarrow$ \emph{dog} for CIFAR-10. For CIFAR-100, the noise flips each class into the next, circularly within super-classes. 

\noindent\textbf{Network and Optimizer.} We use ResNet34 \cite{he2016deep} for both datasets, and train them using SGD with a momentum of 0.9, a weight decay of 0.001, and a batch size of 128. The networks are trained for 300 epochs. We use the cosine annealing learning rate \cite{loshchilov2016sgdr} where the maximum number of epoch for each period is 10, the maximum and minimum learning rate is set to 0.02 and 0.001 respectively. We warm up our networks 10 epochs for CIFAR-10 and 30 epochs for CIFAR-100 with cross-entropy loss. We do not perform early stopping since we don’t assume the presence of clean validation data. All test accuracies
are recorded from the last epoch of training. The reason that we train the model 300 epochs is to fully evaluate whether the model will memorize the mislabeled samples, avoiding the interference caused by early stopping \cite{li2020gradient}. 

\noindent\textbf{Hyperparameters.} We set $K=1$ since we find that a larger $K$ worsens the performance. $E_{s}$ should be smaller than the epoch that model starts memorizing mislabeled samples. We estimate this epoch by \cite{lu2022selc}. In this paper, we fix $E_{s}=60$ and $\alpha=0.9$ by default. More discussions on choice of $K$ and $\alpha$ can be found in Section \ref{sec:choiceOfK}.


\begin{table*}
	\caption{Test Accuracy (\%) on CIFAR-10 and CIFAR-100 with different ratios of symmetric and asymmetric label noise. We compare with existing methods under the same backbone ResNet34 \cite{he2016deep}. The average accuracy and standard deviation of 3 random runs are reported. symm/asymm represent symmetric/asymmetric label noise respectively. \textbf{Bold} indicates the best results. - indicates the result is not reported.} 
	\begin{center}
		\resizebox{1.0\textwidth}{!}{
			\centering
			\begin{tabular}{ |p{25mm} |c c c  c cc c c c c c| } 
				\hline
				\multicolumn{2}{|c}{\multirow{1}{*}{Dataset}} & \multicolumn{5}{c}{CIFAR-10} &\multicolumn{5}{c|}{CIFAR-100} \\ 
				\multicolumn{2}{|c}{\multirow{1}{*}{Noise type}} & \multicolumn{4}{c}{symm} &\multicolumn{1}{c}{asymm} & \multicolumn{4}{c}{symm}&\multicolumn{1}{c|}{asymm} \\ 
				\cline{3-6}  \cline{8-11}
				\multicolumn{2}{|c}{\multirow{1}{*}{Method/Noise ratio}}& 20\% & \multicolumn{1}{c}{40\%}&\multicolumn{1}{c}{60\%}&\multicolumn{1}{c}{80\%}& \multicolumn{1}{c}{40\%} & \multicolumn{1}{c}{20\%}&\multicolumn{1}{c}{40\%}&\multicolumn{1}{c}{60\%}&\multicolumn{1}{c}{80\%}&\multicolumn{1}{c|}{40\%}\\
				\hline
				\multirow{1}{*}{CE} &  & 86.98 $\pm$ 0.12 &81.88 $\pm$ 0.29&74.14 $\pm$ 0.56&53.82 $\pm$ 1.04&80.11 $\pm$ 1.44&58.72 $\pm$ 0.26&48.20 $\pm$ 0.65&37.41 $\pm$ 0.94&18.10 $\pm$ 0.82&42.74 $\pm$ 0.61\\ 
				\multirow{1}{*}{F-correction \cite{patrini2017making}} && 87.99 $\pm$ 0.36 &83.25 $\pm$ 0.38&74.96 $\pm$ 0.65&54.64 $\pm$ 0.44&83.55 $\pm$ 0.58&39.19 $\pm$ 2.61&31.05 $\pm$ 1.44&19.12 $\pm$ 1.95&8.99 $\pm$ 0.58&34.44 $\pm$ 1.93\\ 
				
				\multirow{1}{*}{Bootstrap \cite{reed2014training}} && 86.23 $\pm$ 0.23 &82.23 $\pm$ 0.37&75.12 $\pm$ 0.56&54.12 $\pm$ 1.32&81.21 $\pm$ 1.47&58.27 $\pm$ 0.21&47.66 $\pm$ 0.55&34.68 $\pm$ 1.10&21.64 $\pm$ 0.97&45.12 $\pm$ 0.57\\ 
				
				\multirow{1}{*}{GCE \cite{zhang2018generalized}} & &89.83 $\pm$ 0.20 &87.13 $\pm$ 0.22&82.54 $\pm$ 0.23&64.07 $\pm$ 1.38&76.74 $\pm$ 0.61&66.81 $\pm$ 0.42&61.77 $\pm$ 0.24&53.16 $\pm$ 0.78&29.16 $\pm$ 0.74&47.22 $\pm$ 1.15\\ 
				\multirow{1}{*}{SCE \cite{wang2019symmetric}} & & 89.83 $\pm$ 0.32 &87.13 $\pm$ 0.26&82.81 $\pm$ 0.61&68.12 $\pm$ 0.81&82.51 $\pm$ 0.45&70.38 $\pm$ 0.13&62.27 $\pm$ 0.22&54.82 $\pm$ 0.57&25.91 $\pm$ 0.44&49.32 $\pm$ 0.87\\
				
				
				\multirow{1}{*}{NCE+RCE \cite{ma2020normalized}} & & - &86.02 $\pm$ 0.09&79.78 $\pm$ 0.50&52.71 $\pm$ 1.90&79.59 $\pm$ 0.40&-&59.48 $\pm$ 0.56&47.12 $\pm$ 0.62&25.80 $\pm$ 1.12&46.79 $\pm$ 0.96\\
				
				\multirow{1}{*}{Joint Optim \cite{tanaka2018joint}} & & 92.25 &90.79&86.87&69.16&-&58.15&54.81&47.94&17.18&-\\
				
				\multirow{1}{*}{PENCIL \cite{yi2019probabilistic}} & & - &-&-&-&\textbf{91.01}&-&69.12 $\pm$ 0.62&57.79 $\pm$ 3.86&fail&63.61 $\pm$ 0.23\\
				
				\multirow{1}{*}{RoG+D2L \cite{lee2019robust}} & & - &87.00&78.00&-&-&-&64.90&40.60&-&-\\
				
				\multirow{1}{*}{M-correction \cite{pmlr-v97-arazo19a}}& &-& 92.30 &86.10&74.10&-&-&70.10&59.50&39.50&-\\
				
				\multirow{1}{*}{MentorNet \cite{jiang2017mentornet}}& &92.00& 91.20 &74.20&60.00&-&73.50&68.50&61.20&35.50&-\\

				\multirow{1}{*}{O2U-net \cite{huang2019o2u}} & & - &90.30&-&43.40&-&-&69.20&-&39.40&-\\
				\multirow{1}{*}{NLNL \cite{kim2019nlnl}} & & \textbf{94.23} &92.43&88.32&-&89.86&71.52&66.39&56.51&-&45.70\\
				
				\multirow{1}{*}{DAC \cite{thulasidasan2019combating}} & & 92.91 &90.71&86.30&74.84&-&73.55&66.92&57.17&32.16&-\\
				\multirow{1}{*}{SELF \cite{nguyen2020self}} & & - &91.13&-&63.59&-&-&66.71&-&35.56&-\\
				\multirow{1}{*}{ELR \cite{liu2020early}} & & 91.16 $\pm$ 0.08 &89.15 $\pm$ 0.17&86.12 $\pm$ 0.49& 73.86 $\pm$ 0.61&90.12 $\pm$ 0.47&74.21 $\pm$ 0.22&68.28 $\pm$ 0.31& 59.28 $\pm$ 0.67&29.78 $\pm$ 0.56&\textbf{73.26 $\pm$ 0.64}\\
				
				
				\hline
				\multirow{1}{*}{MixNN (Ours)}& &93.91 $\pm$ 0.12&\textbf{92.89} $\pm$ \textbf{0.02}&\textbf{91.66} $\pm$ \textbf{0.07} &\textbf{86.08} $\pm$ \textbf{1.01}& 90.25 $\pm$ 0.76& \textbf{74.81} $\pm$ \textbf{0.14}&\textbf{72.97} $\pm$ \textbf{0.14}&\textbf{67.56} $\pm$ \textbf{0.17}&\textbf{48.81} $\pm$ \textbf{0.06}&68.18 $\pm$ 0.11 \\
				\hline

			\end{tabular}
		}
	\end{center}
	\label{table:cifar10and100_resnet34}
\end{table*}

\noindent\textbf{Results on CIFAR-10 and CIFAR-100} Table \ref{table:cifar10and100_resnet34} shows the classification test accuracies of our approach on CIFAR-10 and CIFAR-100 with different levels of symmetric and asymmetric label noise. As we can see, MixNN achieves the highest accuracy in most cases, especially in challenging ones. For example, on CIFAR-10 with 80\% symmetric label noise, MixNN outperforms the best state-of-the-art method (74.84\% of DAC) by more than 11\%. In the hardest case (i.e. CIFAR-100 with 80\% symmetric label noise), we observe that most existing methods achieve relatively low test accuracies and PENCIL even fails to converge. However, MixNN still achieves the best accuracy up to 48.81\%. Note that on CIFAR-10/CIFAR-100 with 20\% symmetric label noise, NLNL and DAC obtain superior performance and even outperform our approach. However, these two methods are more complex and perform multiple training stages for different purposes. For example, NLNL performs three training stages including a) Division of training data into either clean or noisy data with a DNN model. b) Training initialized DNN with clean data from the first stage and then updating noisy data’s label following the output of DNN trained with clean data. c) Clean data and label-updated noisy data are both used for training initialized DNN in the final stage. In contrast, our method conducts an end-to-end learning manner which is much simpler than NLNL and also achieves excellent performance. In summary, MixNN shows a consistently strong performance across all datasets with different types and ratios of simulated label noise. 

\subsection{Effectiveness on Large-scale Real-world Datasets}
We evaluate the performance of our approach under the following two large-scale datasets.
\begin{itemize}
	\item \emph{Clothing1M} \cite{xiao2015learning} contains 1 million images of clothing obtained from online shopping websites with 14 classes, such as T-shirt, sweater, and so on. The labels are generated by using the surrounding texts of the images that are provided by the sellers and thus contain many wrong labels. The label noise rate is around 38.46\%, with some pairs of classes frequently confused with each other (e.g. knitwear and sweater). 
	\item \emph{Webvision} \cite{li2017webvision} is a large web images dataset that contains more than 2.4 millions of images crawled from the Flickr and Google Images Search. The label noise level of Webvision is estimated at 20\%. Following \cite{chen2019understanding}, we compare the baseline methods on the first 50 classes of Google image subset. We not only test the trained model on the human-annotated WebVision validation set, but also test the model on ILSVRC12 validation set to evaluate generalization capability. 
\end{itemize}

\noindent\textbf{Data Preprocessing} We apply normalization and regular data augmentations (i.e. random crop and horizontal flip) on the training sets. Since real-world images are of different sizes, we perform the cropping consistent with the existing work \cite{chen2019understanding}. Specifically, 224 $\times$ 224 for Clothing1M (after resizing to 256 $\times$ 256), and 227 $\times$ 227 for Webvision.

\noindent\textbf{Network and Optimizer} We use the ResNet-50 \cite{he2016deep} pretrained on ImageNet for Clothing1M. We train the model with batch size 64. The optimization is done using SGD with a momentum of 0.9, and weight decay of 0.001. We use the same cosine annealing learning rate as CIFAR-10 except the minimum learning rate is set to 0.0001 and the total epoch is 200. For each epoch, we randomly sample 2000 mini-batches from the training data ensuring that the classes of the noisy labels are balanced. For Webvision, we use InceptionResNetV2 \cite{szegedy2017inception} as the backbone architecture. All other optimization details are the same as for CIFAR-10, except for the weight decay (0.0005), the batch size (32) and maximum learning rate (0.01).

\noindent\textbf{Results on Clothing1M and Webvision} Table \ref{table:clothing1m} shows the results on Clothing1M dataset. MixNN consistently outperforms other baselines, slightly superior to Joint-Optim. Table \ref{table:webvision} compares MixNN to state-of-the-art methods trained on the Webvision and evaluated on both the WebVision and ImageNet ILSVRC12 validation sets. On WebVision validation set, MixNN produces superior results in terms of top 1 and top 5 accuracies, demonstrating our method is effective on datasets containing real-world label noise. Furthermore, MixNN outperforms other methods in the ILSVRC12 validation set, demonstrating MixNN's excellent generalization capability.

\begin{table*}
	\caption{Comparison with state-of-the-art methods trained on Clothing1M. Results of other methods are taken from original papers. All methods use a ResNet-50 architecture pretrained on ImageNet. }
	\begin{center}
		\resizebox{1.0\textwidth}{!}{
			\centering
			\begin{tabular}{ | c | c c  c c c c c c c|  c |} 
				\hline
				\multicolumn{1}{|c}{Methods}&\multicolumn{1}{|c}{CE}&\multicolumn{1}{c}{F-correction \cite{patrini2017making}}&\multicolumn{1}{c}{GCE \cite{zhang2018generalized}}
				&\multicolumn{1}{c}{Co-teaching \cite{han2018co}}&\multicolumn{1}{c}{SEAL \cite{chen2021beyond}}
				&\multicolumn{1}{c}{SCE \cite{wang2019symmetric}}&\multicolumn{1}{c}{LRT \cite{zheng2020error} }&\multicolumn{1}{c}{Joint-Optim \cite{tanaka2018joint}}&\multicolumn{1}{|c|}{MixNN}\\
				\hline
				\multicolumn{1}{|c}{Test accuracy}&\multicolumn{1}{|c}{69.21}&\multicolumn{1}{c}{69.84}&\multicolumn{1}{c}{69.75}&\multicolumn{1}{c}{70.15}&\multicolumn{1}{c}{70.63} &\multicolumn{1}{c}{71.02} &\multicolumn{1}{c}{71.74}& \multicolumn{1}{c}{72.16}&\multicolumn{1}{|c|}{\textbf{72.39}} \\
				\hline
			\end{tabular}
		}
	\end{center}
	\label{table:clothing1m}
	\vspace{-0.8em}
\end{table*}
\begin{table*}[htb!]
	\caption{Comparison with state-of-the-art methods trained on (mini) WebVision. Results of other baseline methods are taken from original papers. All methods use an InceptionResNetV2 architecture. }
	\begin{center}
		\resizebox{1.0\textwidth}{!}{
			\centering
			\begin{tabular}{ p{25mm} c c c  c c c c c c c| c} 
				\hline
				\multicolumn{3}{|c}{\multirow{1}{*}{}} &\multicolumn{1}{c}{F-correction \cite{patrini2017making}}&\multicolumn{1}{c}{Decoupling \cite{malach2017decoupling}}& \multicolumn{1}{c}{D2L \cite{ma2018dimensionality}}&\multicolumn{1}{c}{MentorNet \cite{jiang2017mentornet}}&\multicolumn{1}{c}{Co-teaching \cite{han2018co}} &\multicolumn{1}{c}{Iterative-CV \cite{chen2019understanding}}&\multicolumn{1}{c}{Crust \cite{mirzasoleiman2020coresets}}&\multicolumn{1}{c}{ODD \cite{song2020robust}}&\multicolumn{1}{|c|}{MixNN}\\
				\hline
				\multicolumn{2}{|c}{\multirow{2}{*}{WebVision}} & \multicolumn{1}{c}{top1}&\multicolumn{1}{c}{61.12}&\multicolumn{1}{c}{62.54}&\multicolumn{1}{c}{62.68}&\multicolumn{1}{c}{63.00}&\multicolumn{1}{c}{63.58} &\multicolumn{1}{c}{65.24} & \multicolumn{1}{c}{72.40}&\multicolumn{1}{c}{74.60}&\multicolumn{1}{|c|}{\textbf{75.39}} \\ 
				\multicolumn{2}{|c}{}& top5 & \multicolumn{1}{c}{82.68}&\multicolumn{1}{c}{84.74}&\multicolumn{1}{c}{84.00}&\multicolumn{1}{c}{81.40}&\multicolumn{1}{c}{85.20}& 85.34 & \multicolumn{1}{c}{89.56}&\multicolumn{1}{c}{90.60}&\multicolumn{1}{|c|}{\textbf{90.71}}\\
				\hline
				\multicolumn{2}{|c}{\multirow{2}{*}{ILSVRC12}} & \multicolumn{1}{c}{top1}&\multicolumn{1}{c}{57.36}&\multicolumn{1}{c}{58.26}&\multicolumn{1}{c}{57.80}&\multicolumn{1}{c}{57.80}&\multicolumn{1}{c}{61.48} &\multicolumn{1}{c}{61.60} & \multicolumn{1}{c}{67.36}&\multicolumn{1}{c}{66.70}&\multicolumn{1}{|c|}{\textbf{70.54}} \\ 
				\multicolumn{2}{|c}{}& top5 & \multicolumn{1}{c}{82.36}&\multicolumn{1}{c}{82.26}&\multicolumn{1}{c}{81.36}&\multicolumn{1}{c}{79.92}&\multicolumn{1}{c}{84.70}& 84.98 & \multicolumn{1}{c}{87.84}&\multicolumn{1}{c}{86.30}&\multicolumn{1}{|c|}{\textbf{90.31}}\\
				\hline

			\end{tabular}
		}
	\end{center}
	\label{table:webvision}
\end{table*}

\subsection{Ablation Study}
We study the effect of removing different components to provide insights into what makes MixNN successful. The results are in Table \ref{tab:ablation}. First, we remove the dynamic weight estimation in MixNN. Instead, we average (AVG) the selected samples or use random (RDM) weights from a Beta distribution in mixing function. We observe that merely averaging the selected samples does not perform well in all noisy cases, especially when the noise ratio is large. In comparison, using random weights may result in a surprising performance. For instance, it achieves excellent performance in CIFAR-100 with 40\% asymmetric label noise. Since asymmetric label noise is concentrated inside each class, the estimated weights may fail to capture the clean probability of difficult samples, forcing the mixing function to use an average way. However, using the random weights is likely to appropriately assign the weights for the challenging samples, reducing the impact of asymmetric label noise. We also remove the label correction to see how it affects the performance. Without label correction, the performance suffers, especially when the noise ratio is high. Further, we investigate the performance of MixNN when using random samples instead of using $K$ nearest neighbours. We observe that the performance is marginally poorer in the low ratio of label noise. However, when the training data contains high ratios of noisy labels, the performance suffers a large decline, demonstrating the benefit of using $K$ nearest neighbours in MixNN.
\begin{table*}
	\caption{Ablation study results in terms of test accuracy (\%) on CIFAR-10 and CIFAR-100. } 
	\begin{center}
		\resizebox{0.8\textwidth}{!}{
			\centering
			\begin{tabular}{| p{50mm} c c  c c c c c |} 
				\hline
				\multicolumn{2}{|c}{\multirow{1}{*}{Dataset}} & \multicolumn{3}{c}{CIFAR-10} &\multicolumn{3}{c|}{CIFAR-100}\\ 
				\multicolumn{2}{|c}{\multirow{1}{*}{Noise type}} & \multicolumn{2}{c}{symm} &\multicolumn{1}{c}{asymm} & \multicolumn{2}{c}{symm}&\multicolumn{1}{c|}{asymm} \\  
				\multicolumn{2}{|c}{\multirow{1}{*}{Noise ratio}}&  \multicolumn{1}{c}{40\%}&\multicolumn{1}{c}{80\%}& \multicolumn{1}{c}{40\%} &\multicolumn{1}{c}{40\%}&\multicolumn{1}{c}{80\%}&\multicolumn{1}{c|}{40\%}\\
				\hline
				\multirow{1}{*}{Baseline CE}  & &81.88 $\pm$ 0.29&53.82 $\pm$ 1.04&80.11 $\pm$ 1.44&48.20 $\pm$ 0.65&18.10 $\pm$ 0.82&42.74 $\pm$ 0.61\\
				\multirow{1}{*}{Mixup}  & &89.46 $\pm$ 0.47&66.32 $\pm$ 0.92&81.66 $\pm$ 1.73&58.12 $\pm$ 0.87&18.77 $\pm$ 1.14&49.61 $\pm$ 1.39\\
				\hline
				\multirow{1}{*}{MixNN} & &\textbf{92.89} $\pm$ \textbf{0.02} &\textbf{86.08} $\pm$ \textbf{1.01}&\textbf{90.25} $\pm$ \textbf{0.76}&\textbf{72.97} $\pm$ \textbf{0.14}&\textbf{48.81} $\pm$ \textbf{0.06}&68.18 $\pm$ 0.11\\
				\multirow{1}{*}{MixNN w/o weight estimation (AVG)}  & &88.79 $\pm$ 0.33&58.53 $\pm$ 0.15&82.27 $\pm$ 0.10&64.21 $\pm$ 0.19&27.91 $\pm$ 0.31&59.73 $\pm$ 0.21\\
				\multirow{1}{*}{MixNN w/o weight estimation (RDM)}  & &91.92 $\pm$ 0.09&78.72 $\pm$ 0.13&89.91 $\pm$ 0.27&71.08 $\pm$ 0.25&41.14 $\pm$ 0.18&\textbf{72.98} $\pm$ \textbf{0.47}\\
				\multirow{1}{*}{MixNN w/o label correction} & &91.73 $\pm$ 0.01&74.31 $\pm$ 0.49&88.73 $\pm$ 0.11&66.09 $\pm$ 0.50&30.18 $\pm$ 0.52&63.66 $\pm$ 0.43\\
				\multirow{1}{*}{MixNN w/o $K$ nearest neighbours}  & &92.19 $\pm$ 0.04&76.58 $\pm$ 0.25&88.69 $\pm$ 0.17&71.46 $\pm$ 0.08&42.02 $\pm$ 0.48&68.61 $\pm$ 0.16\\
				
				\hline

			\end{tabular}
		}
	\end{center}
	\label{tab:ablation}
\end{table*}

\begin{figure*}
	\centering
	\subfigure[Accuracy.]{
		\includegraphics[width=0.48\columnwidth]{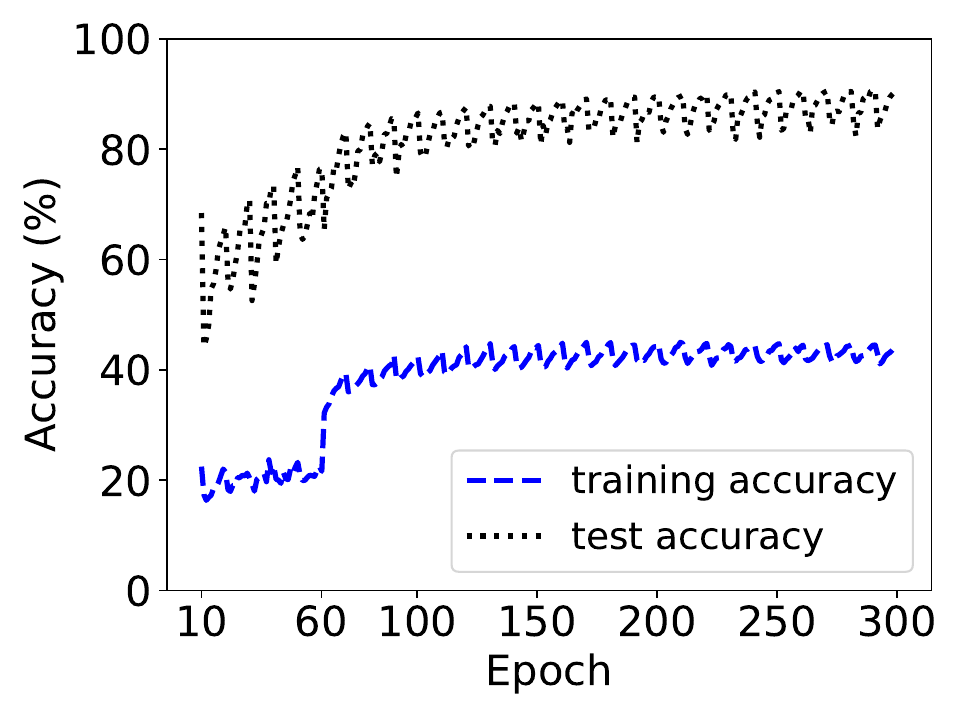}
		\label{fig:accuracy_mixnn}
	}
	\subfigure[Gradient Coefficient.]{
		\includegraphics[width=0.48\columnwidth]{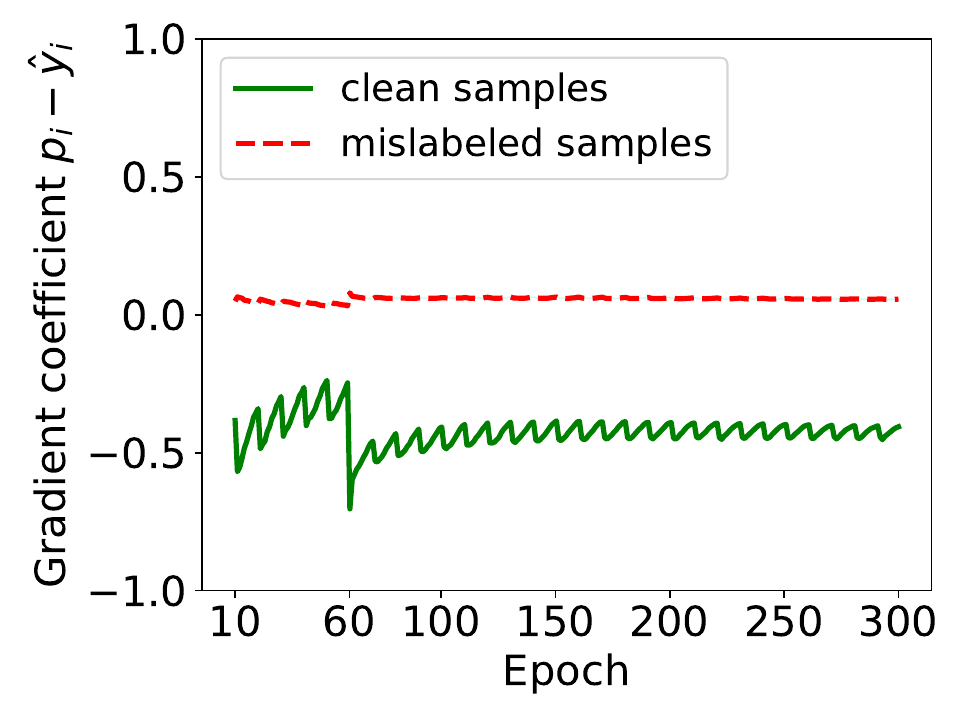}
		\label{fig:grad_mixnn}
	}
	\subfigure[The influence of $K$.]{
		\label{fig:K}
		\includegraphics[width=0.48\columnwidth]{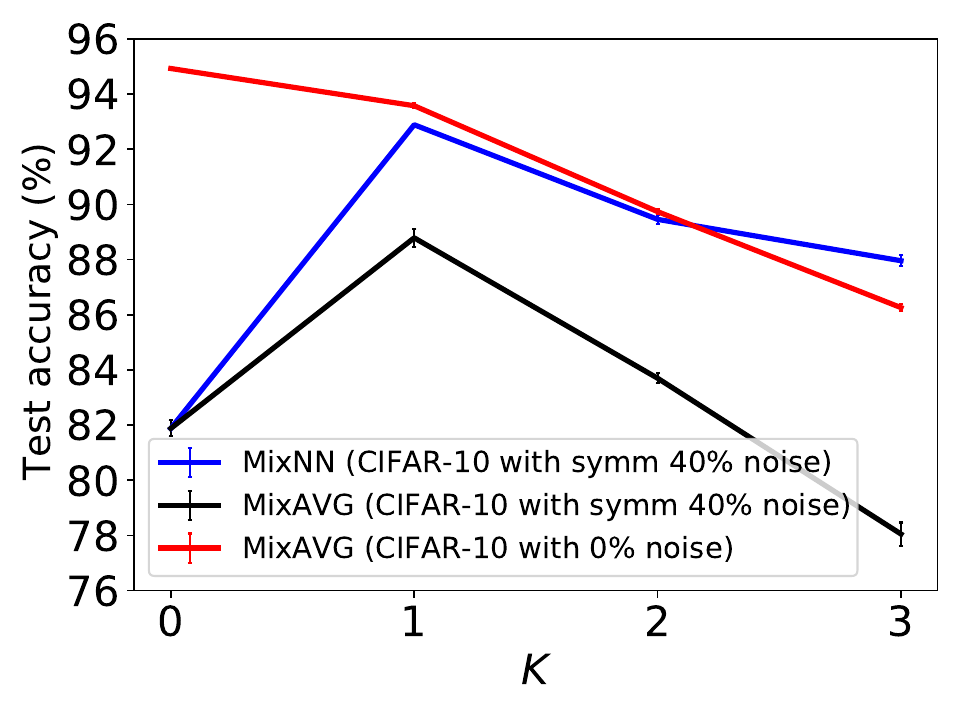}}
	\subfigure[Sensitivity of $\alpha$.]{
		\includegraphics[width=0.48\columnwidth]{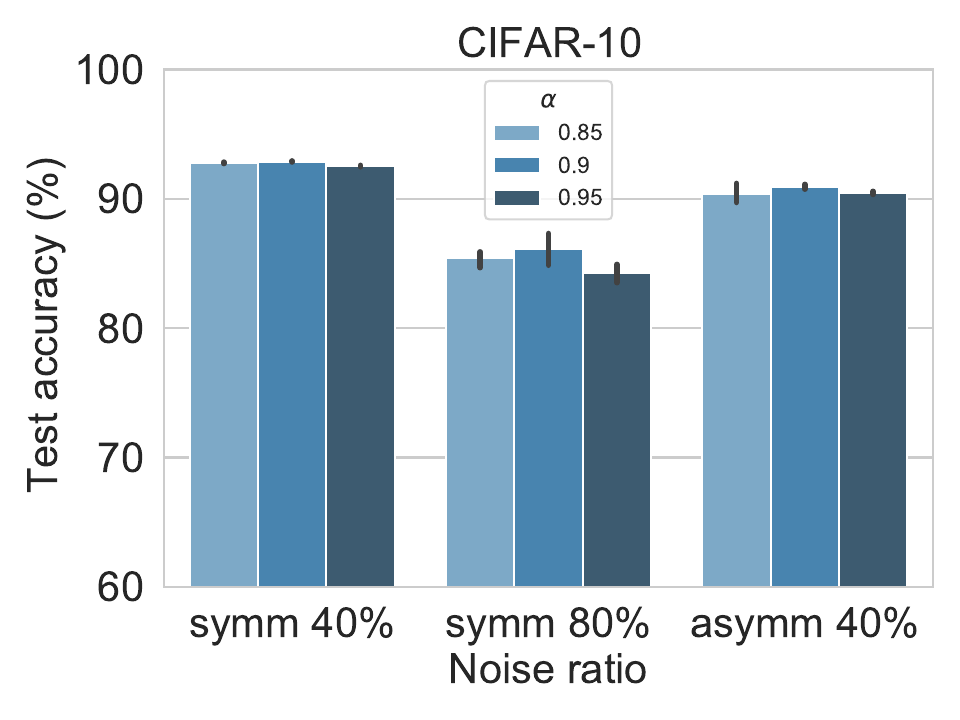}
		\label{fig:alpha}
	}
	\caption{For plots (a) and (b), we train MixNN on CIFAR-10 with 60\% symmetric label noise. Plot (a) shows the training and test accuracy vs. the number of epochs. Plot (b) shows the gradient coefficient vs. the number of epochs. Plot (c) shows the effect of different $K$ on performance. Plot (d) shows the sensitivity of $\alpha$ on performance.  }
	\label{fig:analyses}
\end{figure*}

\begin{figure}
	\centering
	\subfigure[Original]{
		\includegraphics[width=0.48\columnwidth]{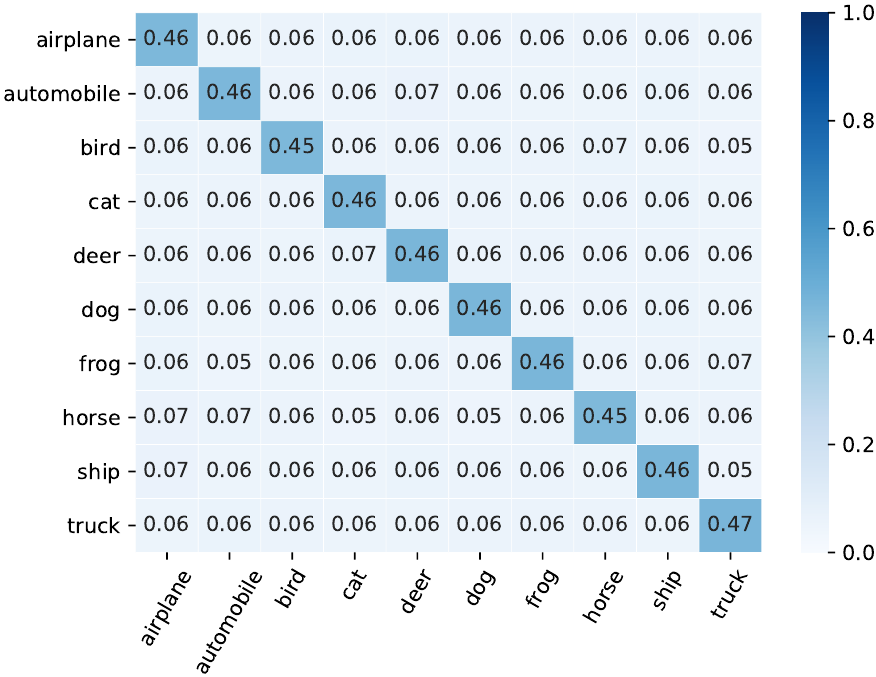}}
	\subfigure[After MixNN]{
		\includegraphics[width=0.48\columnwidth]{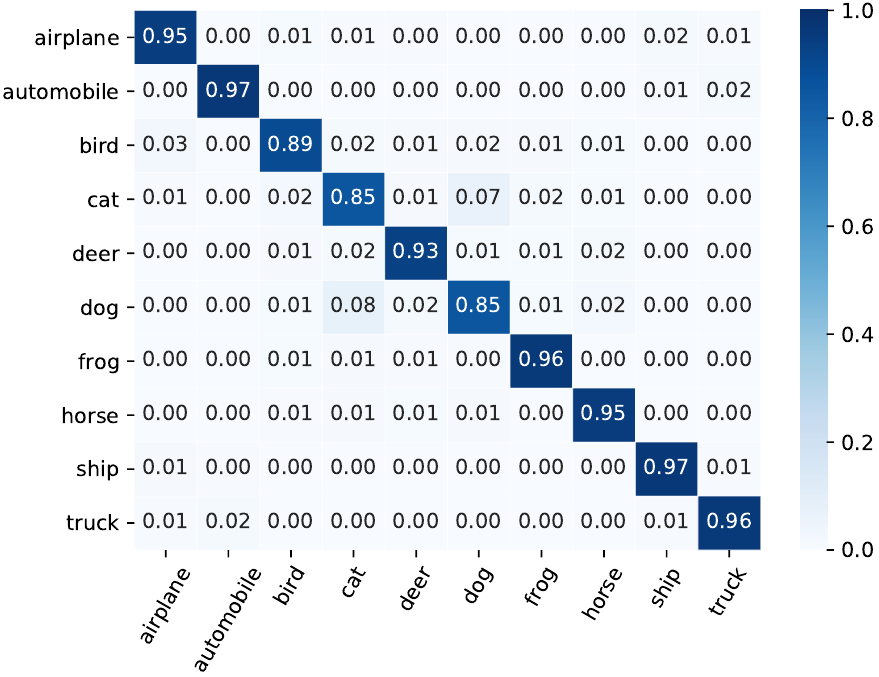}}
	\caption{Confusion matrix of corrected labels w.r.t clean labels on CIFAR-10 with 60\% symmetric label noise. The y-axis is the clean labels. The x-axis is the original noisy labels and corrected labels for plots (a) and (b), respectively. }
	\label{fig:label_correct}
\end{figure}

\subsection{Learning Stability and Gradient Analysis}
In section \ref{sec:pre}, we have demonstrated the failure of CE when trained DNNs with noisy labels. The training accuracy constantly increases indicates the DNNs eventually overfit the noisy labels, resulting in a drop of accuracy on the clean test set. To verify the denoising effect of MixNN, we show its learning stability by plotting the training and test accuracy vs. the number of epochs in Fig. \ref{fig:accuracy_mixnn} on the CIFAR-10 with 60\% label noise. We observe that the training accuracy stabilizes at around 40\% after 60 epochs. It means the DNNs only fit the clean samples, resulting in no drop in test accuracy.
		
%

\begin{figure*}
	\centering
	\subfigure[CE (symm $\varepsilon=0.2$)]{
		\includegraphics[width=0.45\columnwidth]{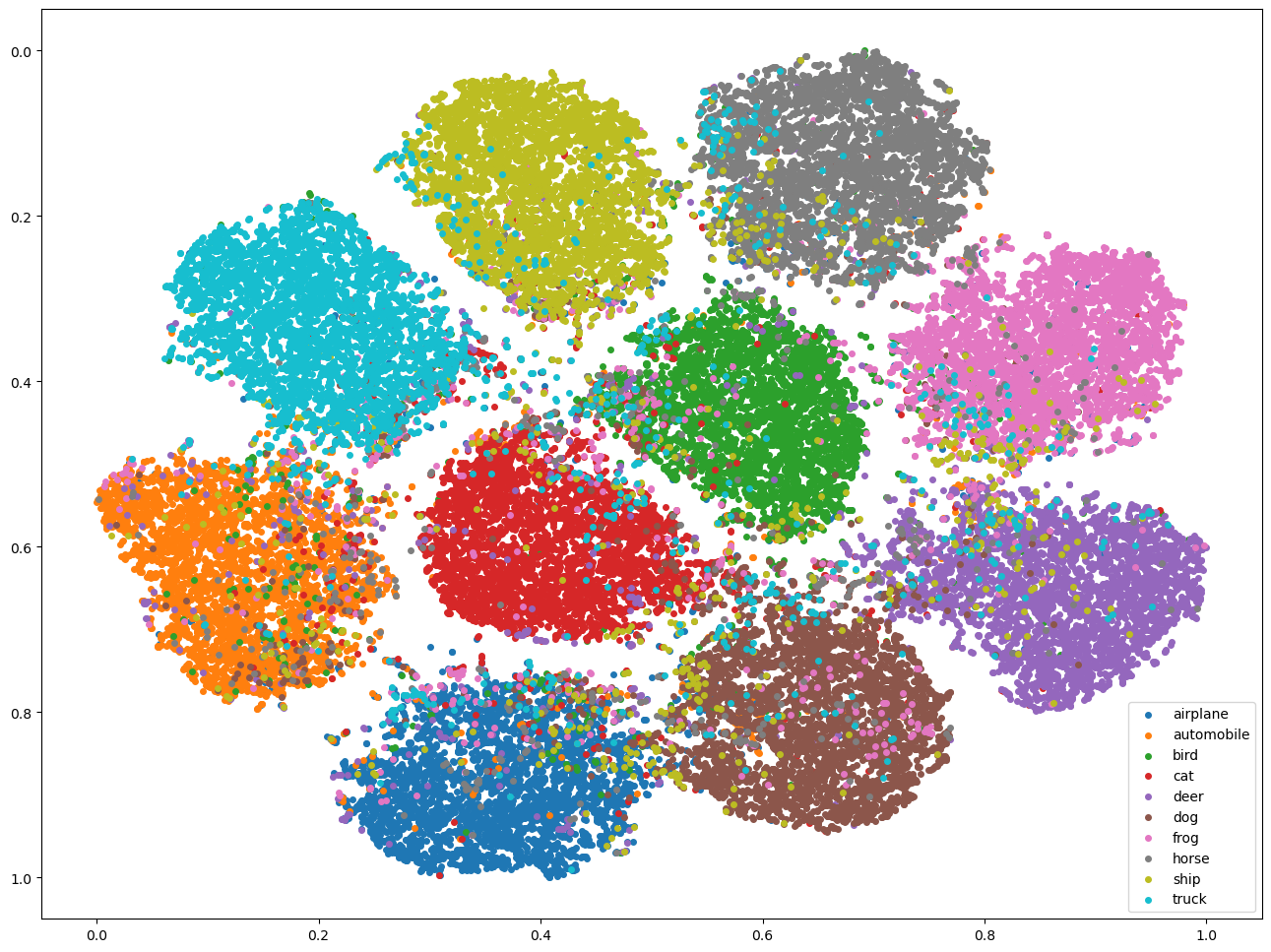}}
	\subfigure[CE (symm $\varepsilon=0.4$)]{
		\includegraphics[width=0.45\columnwidth]{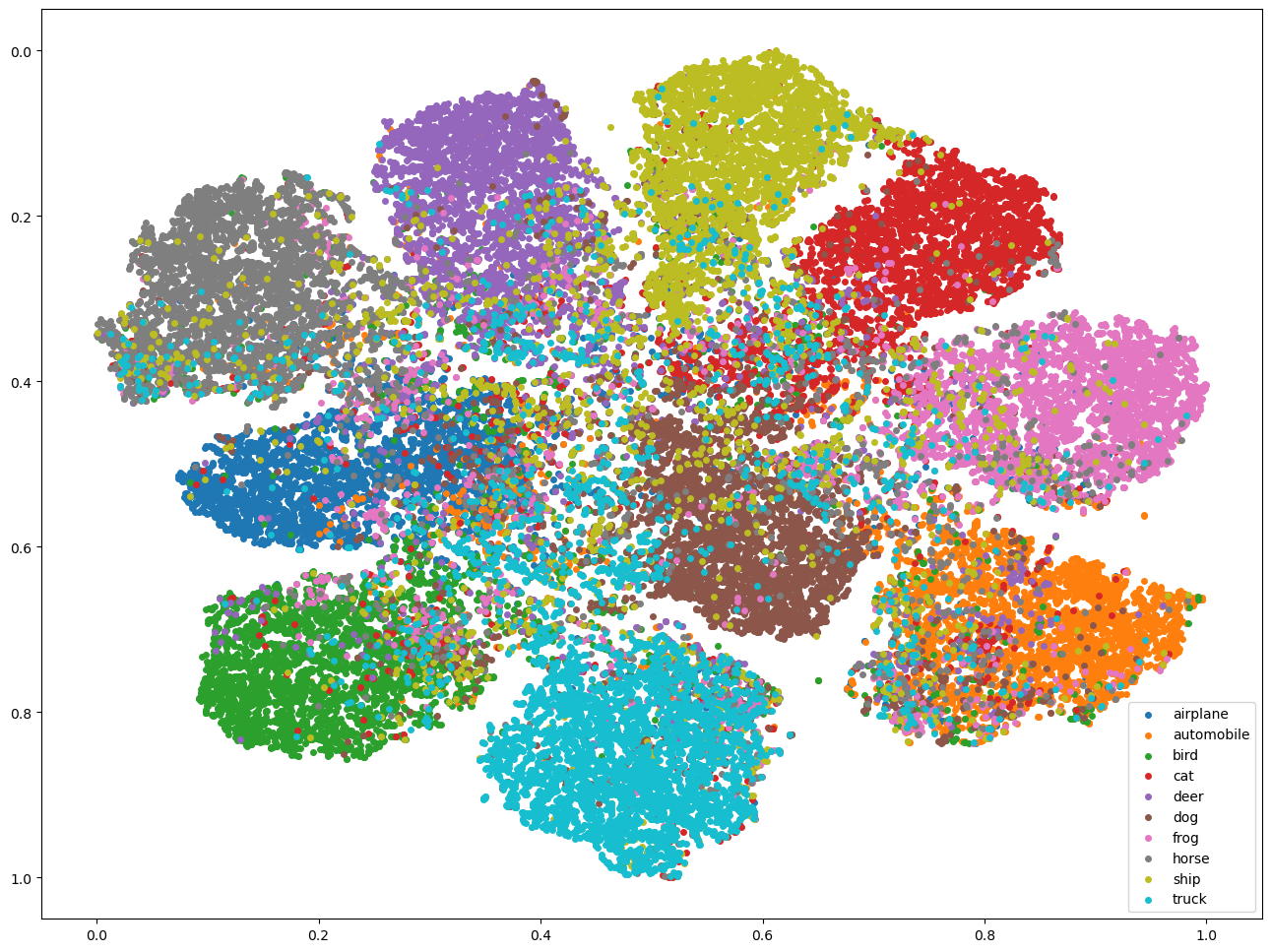}}
	\subfigure[CE (symm $\varepsilon=0.6$)]{
		\includegraphics[width=0.45\columnwidth]{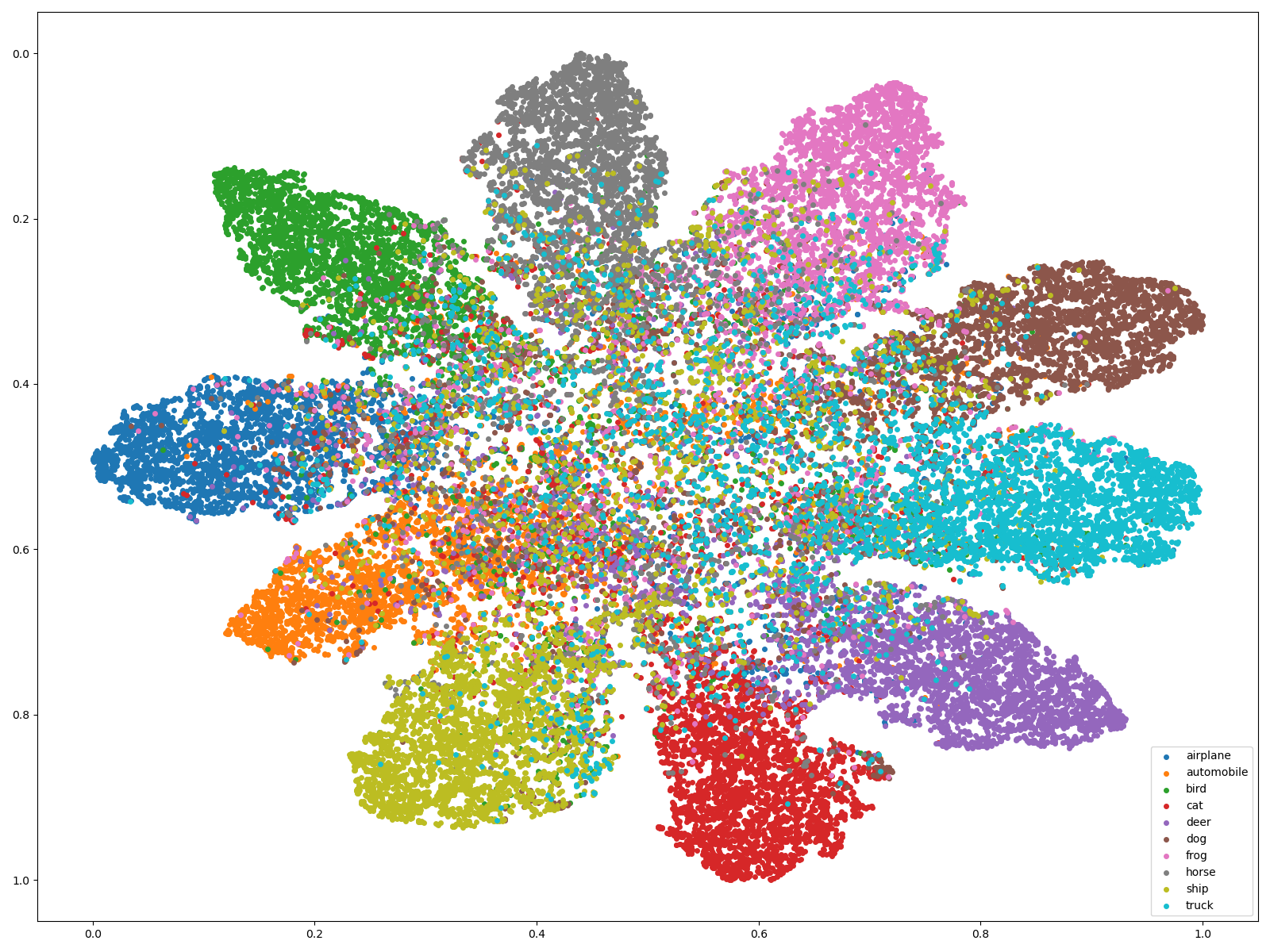}}
	\subfigure[CE (symm $\varepsilon=0.8$)]{
		\includegraphics[width=0.45\columnwidth]{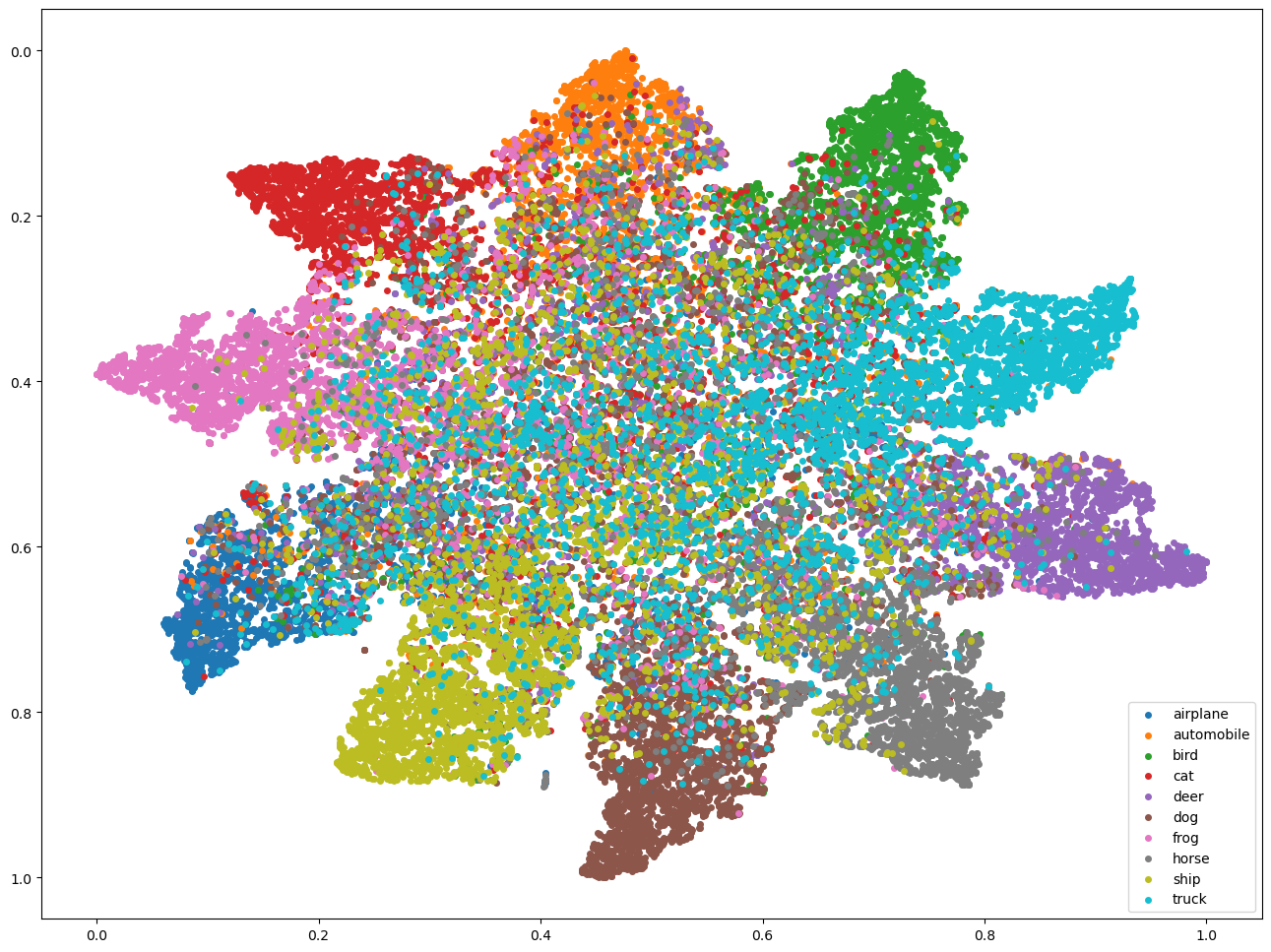}}
	\subfigure[SCE \cite{wang2019symmetric} (symm $\varepsilon=0.2$)]{
		\includegraphics[width=0.45\columnwidth]{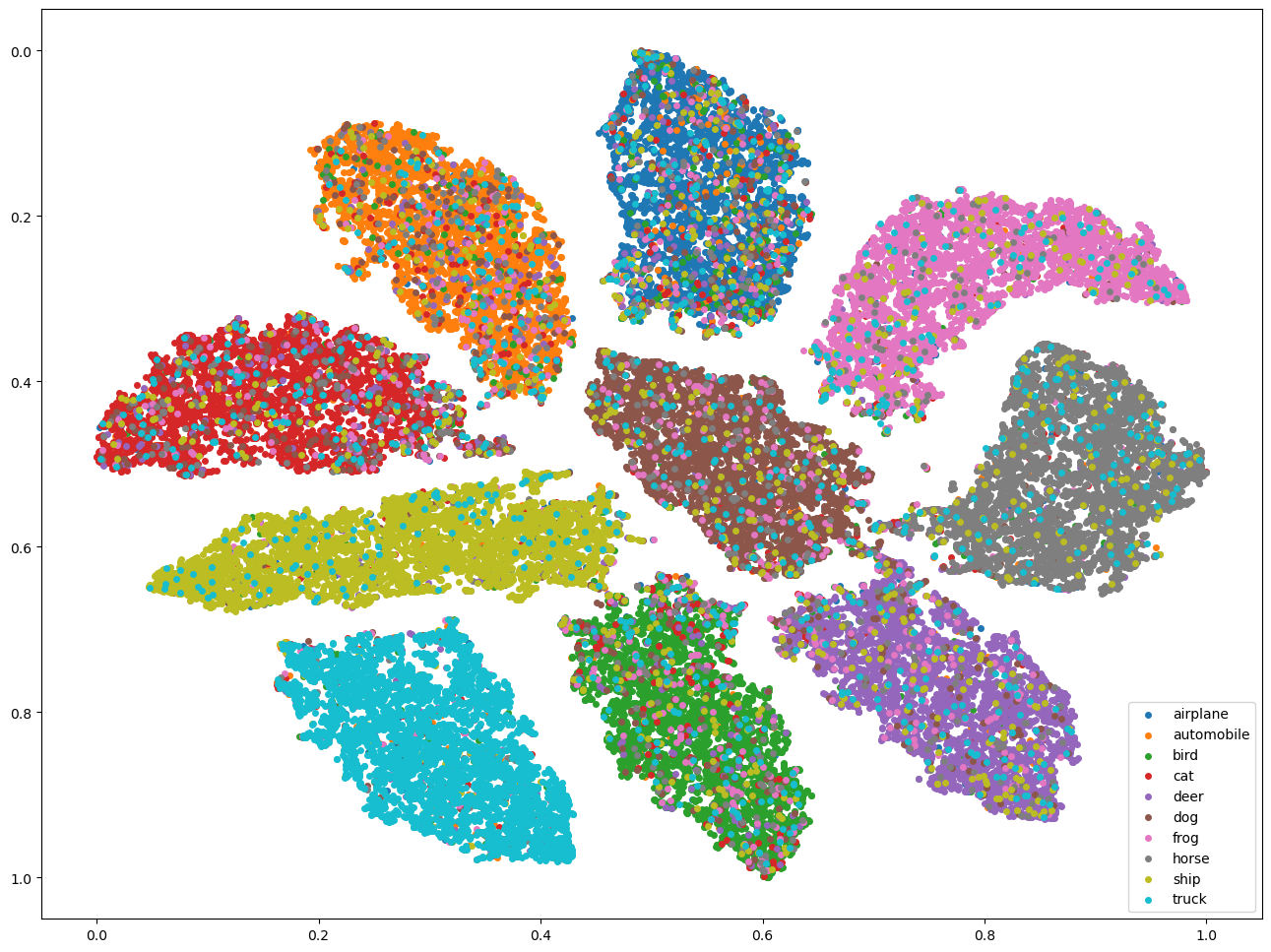}}
	\subfigure[SCE \cite{wang2019symmetric} (symm $\varepsilon=0.4$)]{
		\includegraphics[width=0.45\columnwidth]{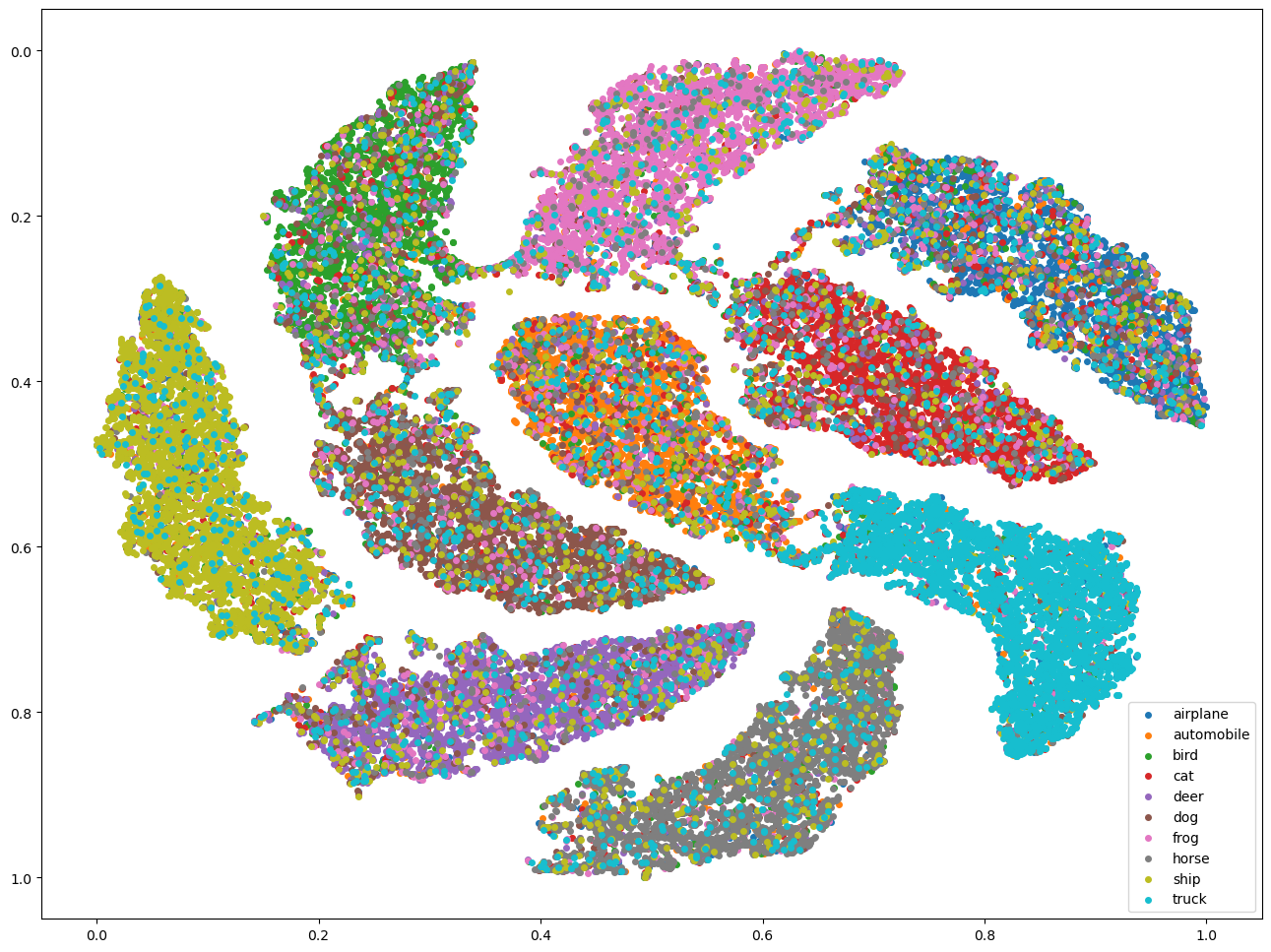}}
	\subfigure[SCE \cite{wang2019symmetric} (symm $\varepsilon=0.6$)]{
		\includegraphics[width=0.45\columnwidth]{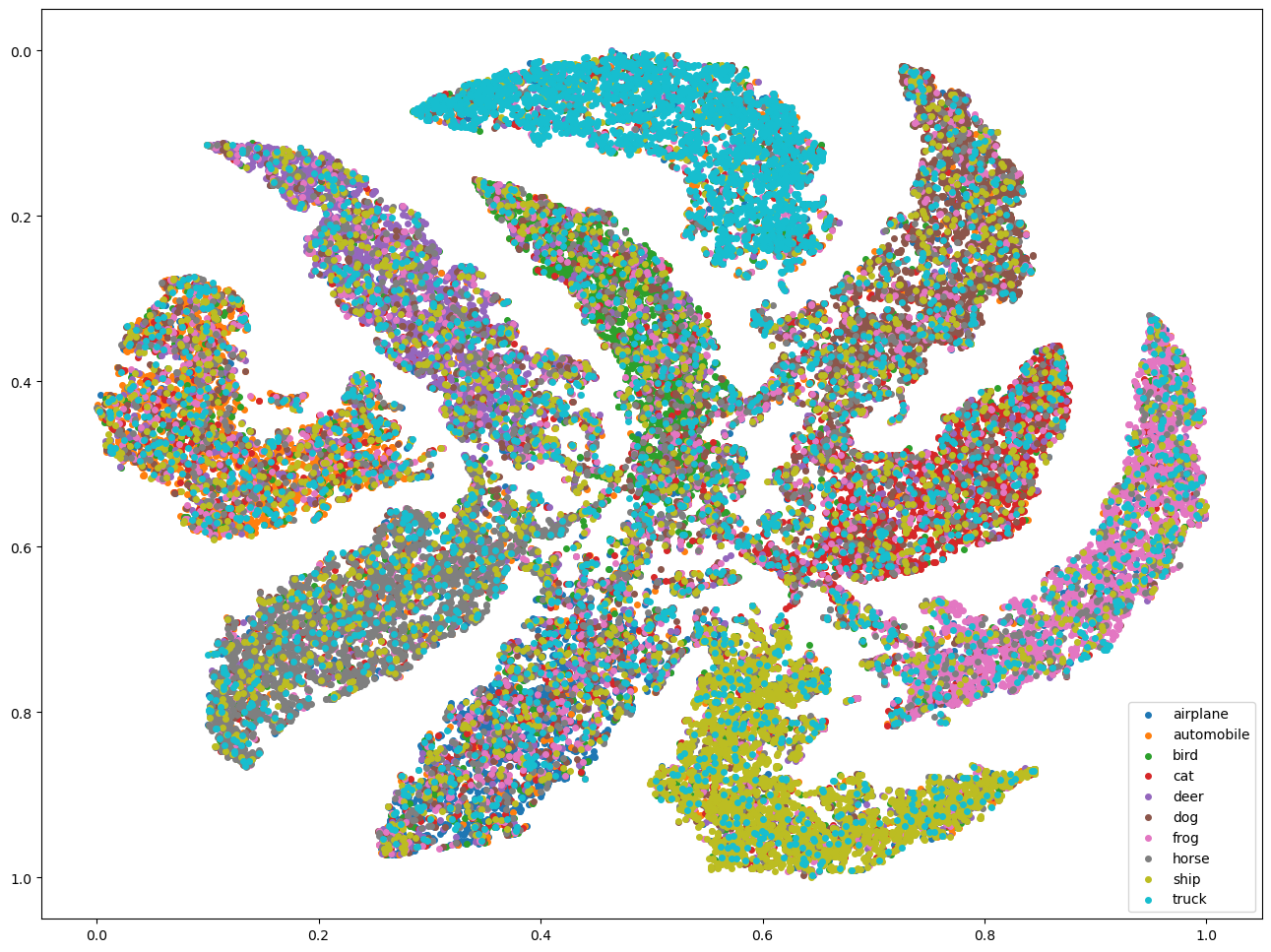}}
	\subfigure[SCE \cite{wang2019symmetric} (symm $\varepsilon=0.8$)]{
		\includegraphics[width=0.45\columnwidth]{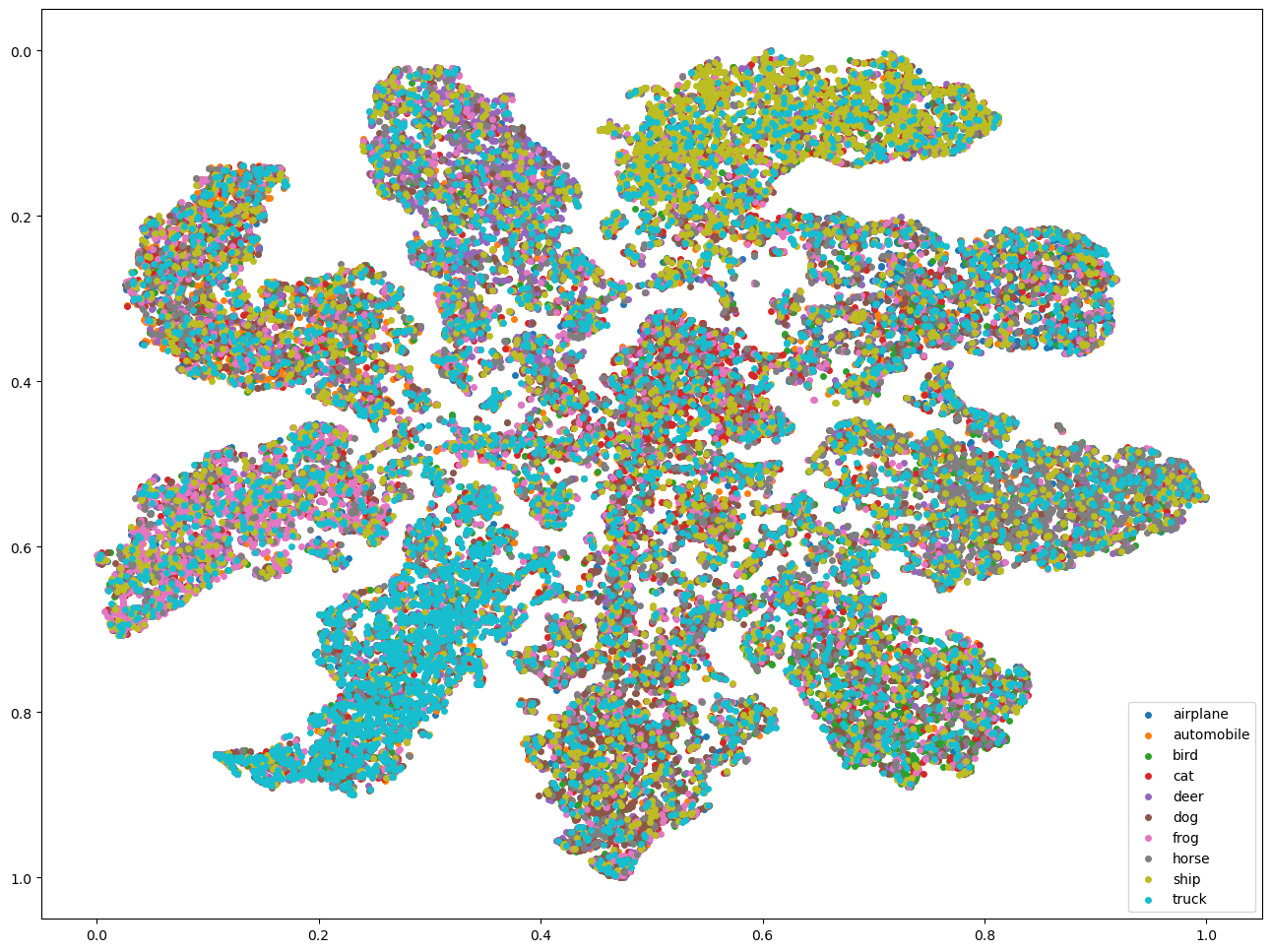}}
	\subfigure[MixNN (symm $\varepsilon=0.2$)]{
		\includegraphics[width=0.45\columnwidth]{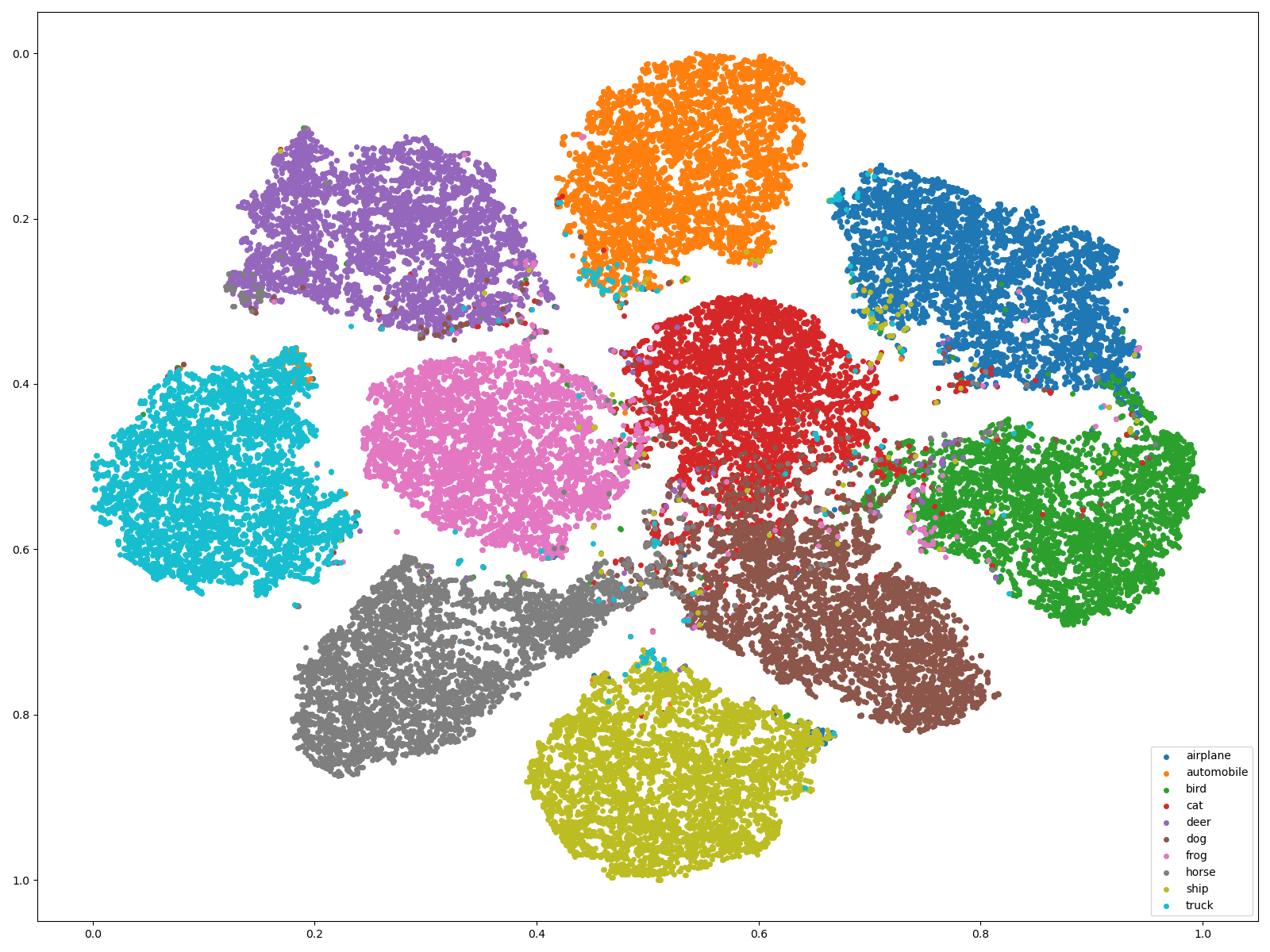}}
	\subfigure[MixNN (symm $\varepsilon=0.4$)]{
		\includegraphics[width=0.45\columnwidth]{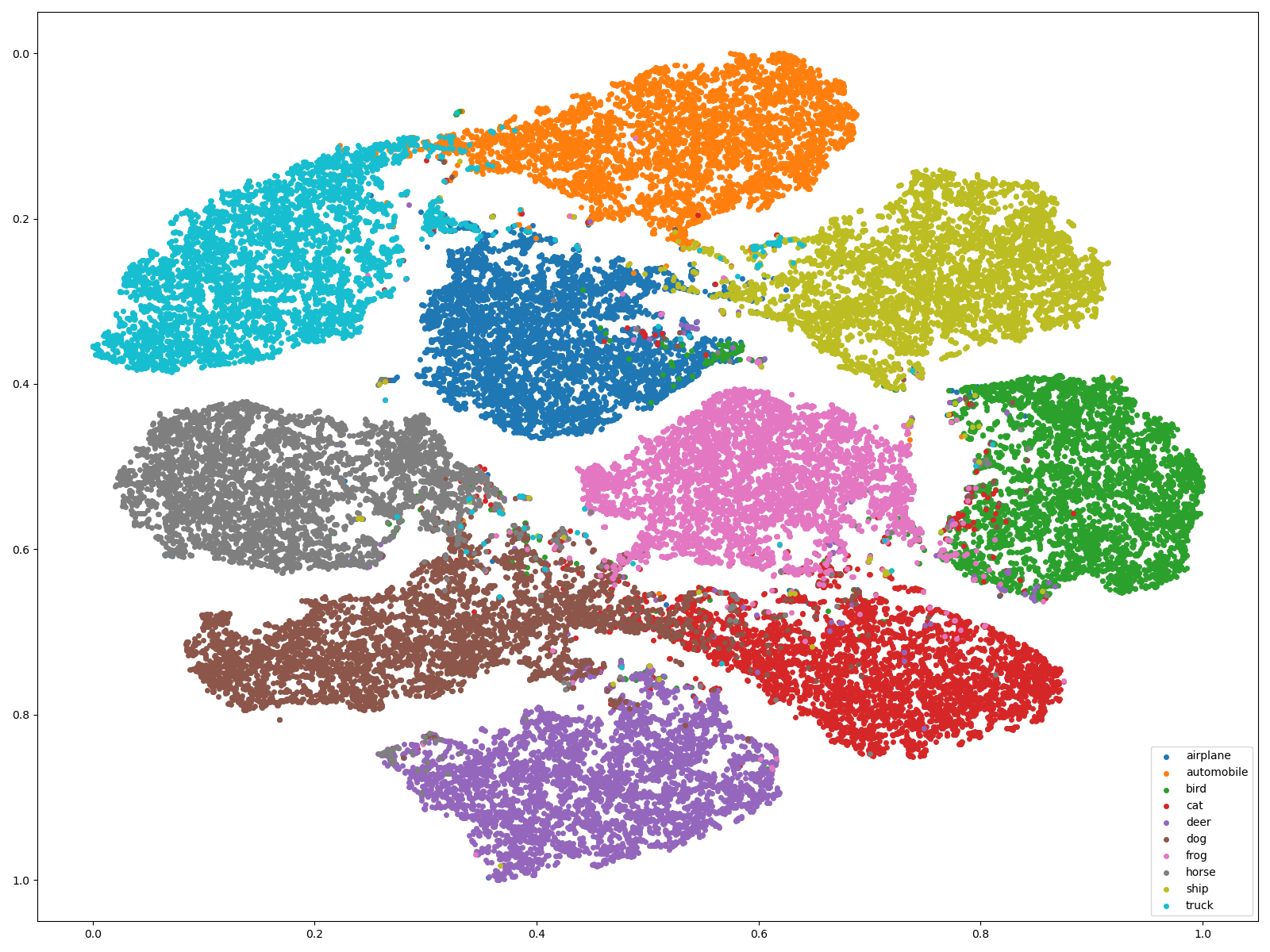}}
	\subfigure[MixNN (symm $\varepsilon=0.6$)]{
		\includegraphics[width=0.45\columnwidth]{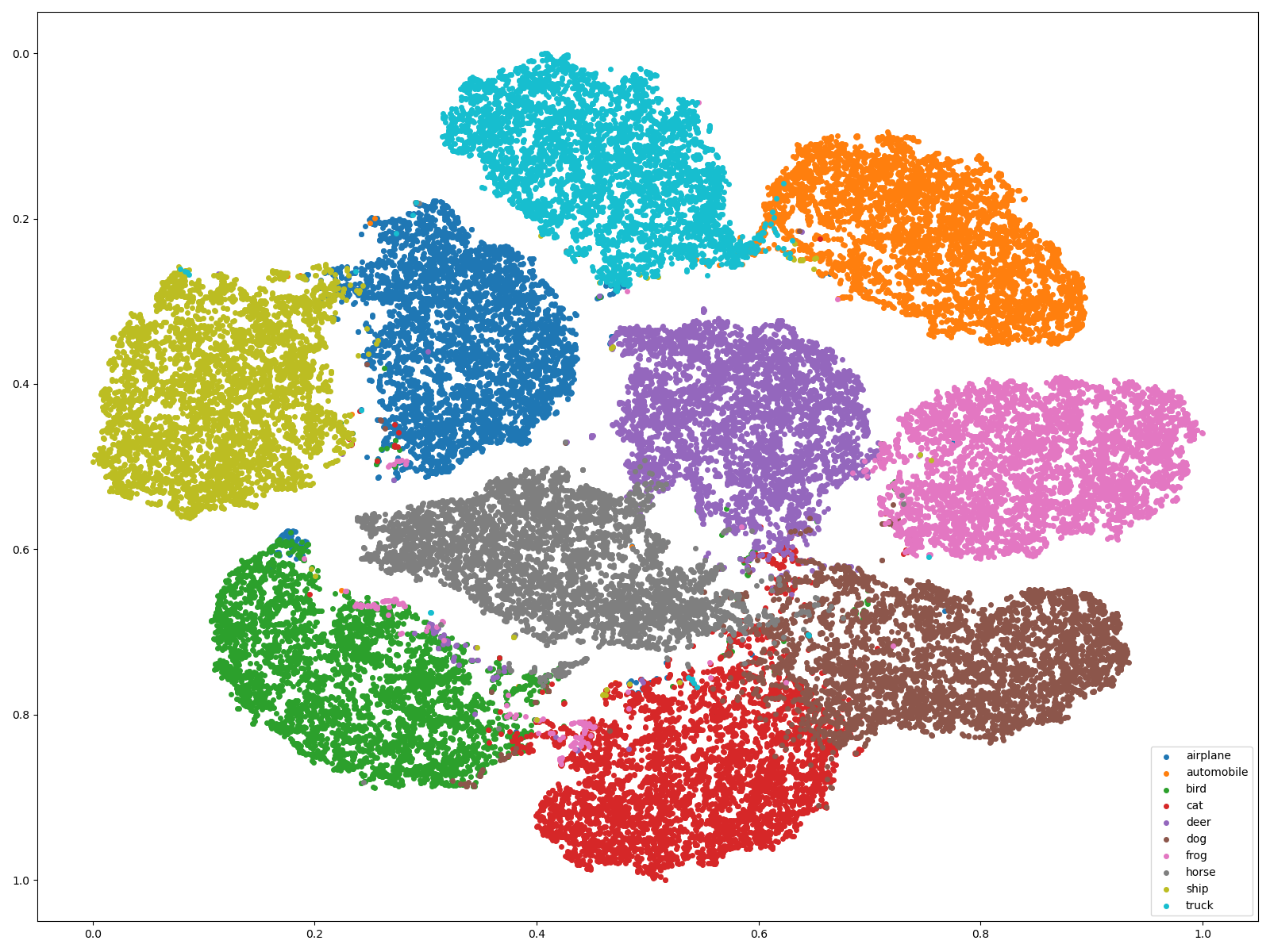}}
	\subfigure[MixNN (symm $\varepsilon=0.8$)]{
		\includegraphics[width=0.45\columnwidth]{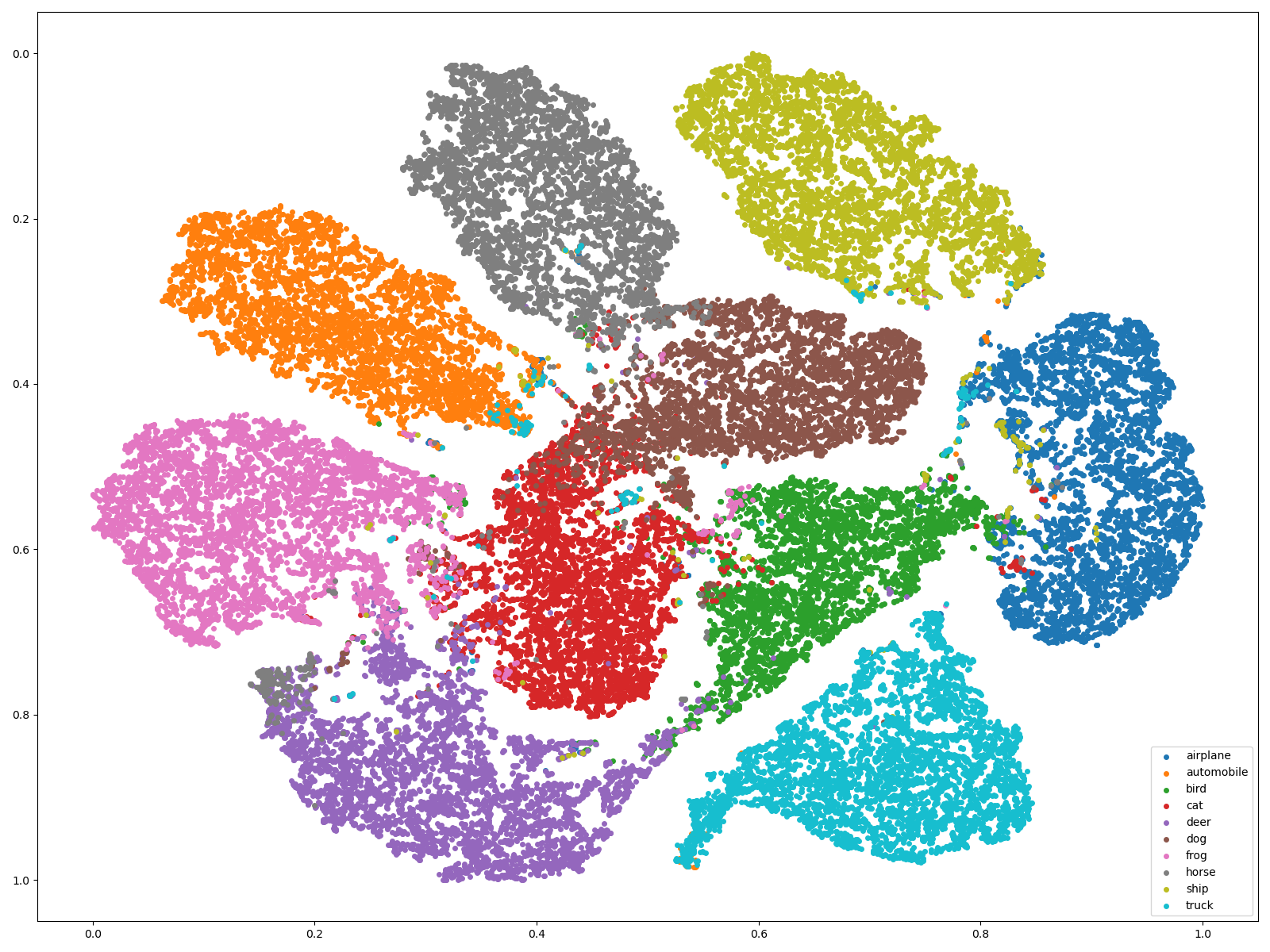}}
	\caption{t-SNE plots of feature representations learned by CE, SCE, and our proposed method MixNN on CIFAR-10 with different ratios of label noise. Different colours represent the different classes in CIFAR-10.}
	\label{fig:representations}
\end{figure*}

We further investigate the gradient coefficient of clean and mislabeled samples when using MixNN. In Fig. \ref{fig:grad_mixnn}. We can observe the differences compared to the CE gradient coefficient in Fig. \ref{fig:memorization}. The gradient of mislabeled samples is close to 0, letting the gradient of clean samples dominates the whole gradient throughout the training. Therefore, MixNN forces the model to learn from clean samples rather than mislabeled samples. In addition, we notice that there is an obvious gradient drop for clean samples when starting using noisy label correction (Note that we start label correction from epoch 60), which demonstrates the label correction effectively suppresses gradient inversion as mentioned in Section \ref{sec:grad_inversion}.
\subsection{Choice of $K$ and $\alpha$}
\label{sec:choiceOfK}
The mixing functions Eq. \ref{eq:mix_x} and Eq. \ref{eq:mix_y} contain a parameter $K$ to control the number of nearest neighbours mixed with original sample. To study the effect of $K$ on our approach, we test different $K$ on CIFAR-10. As shown in Fig. \ref{fig:K}, MixNN achieves the worst performance as it reduces to regular training (i.e. CE) when $K=0$. However, the performance of MixNN starts decreasing with the continuing increase of $K$. To explore whether the performance drop is due to dynamic weight estimation, we add another set of experiments where we simply use averaging strategy (abbreviated as MixAVG). We observe that MixAVG has a similar tendency on performance with the increase of $K$. Therefore, dynamic weight estimation is not the reason causing performance drop. We then conjecture that the reason is the over-mixture of input images. Since the resulting image is the mixture of $K$ images, larger $K$ leads to a more complex input. To avoid the influence of noisy labels, we conduct another set of experiments that use MixAVG on clean training data. We observe a similar performance drop when increasing $K$. Therefore, a reasonable $K$ can suit our needs, while a large $K$ makes it more difficult for the model to learn from complicated inputs, resulting in performance degradation. Figure \ref{fig:alpha} shows the test accuracy when using different $\alpha\in\{0.85,0.9,0.95\}$ on CIFAR-10. We observe that the sensitivity to hyperparameter $\alpha$ is quite mild.



\subsection{Effectiveness of Noisy Label Correction}
\label{sec:correction_accuracy}
In Section \ref{sec:label_correct}, the estimated target $\bm{t}_{i}$ is calculated by an exponential moving average between the given noisy labels and model predictions, thus our method can gradually correct the noisy labels. We use the confusion matrix to evaluate the quality of labels before and after using our method. In Figure \ref{fig:label_correct} shows the results on CIFAR-10 with 60\% symmetric label noise. Our approach corrects the noisy labels impressively well for all classes. Furthermore, our mixing strategy is more reliable to obtain the accurate mixed labels using the corrected soft labels. We also calculate the correction accuracy of our strategy. Correction accuracy can be calculated by $\frac{1}{N}\sum_{i}^{N}\mathbbm{1} \{\mathrm{argmax}\ \bm{y}_{i}=\mathrm{argmax}\ \bm{t}_{i}\}$, where $\bm{y}_{i}$ is the ground truth label of training sample $\bm{x}_{i}$. We evaluate the correction accuracy on CIFAR-10 and CIFAR-100 with 40\% symmetric label noise. Our approach successfully corrects a huge amount of labels and obtains recovered accuracy of 95.25\% and
84.86\%, respectively. More results on other noise ratios can be found in supplementary materials. 



\subsection{Feature Representations}
We further investigate the representations learned by our approach compared to that learned by CE loss and SCE \cite{wang2019symmetric}. We extract the high-dimensional representations at the penultimate layer and project them to a 2D embedding by using t-SNE \cite{van2008visualizing}. The projected representations are illustrated in Fig. \ref{fig:representations} for 20\%, 40\%, 60\% and 80\% symmetric label noise respectively. Under all settings, the feature representations learned by our approach are of significantly better quality than that of CE and SCE with more separated and clearly bounded clusters. We find that SCE always keeps the feature representations of mislabeled samples in their true classes, the same as what CE does in the early learning stage, which prevents the model from memorizing them. However, the boundary formed by the SCE becomes increasingly blurred as the noise ratio rises (see $\varepsilon=0.8$ case). In contrast, our approach gradually corrects the noisy labels to clean labels, resulting in most of the feature representations in different classes are corrected.

\section{Conclusion}
 \label{sec:conclusion}
In this paper, we explore the representation distribution in the early learning stage and propose a method called MixNN for robust learning with noisy labels. MixNN mitigates the negative influence of noisy labels by training with the synthetic samples obtained by mixing the original training samples with their $K$ nearest neighbours. There are two key elements to our approach. The mixing strategy is dynamically adjusted by the learned Gaussian Mixture Model on per-sample loss distribution. The soft targets in our loss function are gradually corrected by using an exponential moving average on the given labels and model predictions. Through extensive experiments across multiple datasets with simulated and real-world label noise, we demonstrate that MixNN consistently exhibits substantial performance improvements compared to state-of-the-art methods. Importantly, the proposed approach works with any classifier “out-of-the-box” without any changes to architecture or training procedure. In the future, we are interested in adapting MixNN to other research domains such as natural language process (NLP) and object detection, and believe MixNN is a promising algorithm for training robust DNNs against label noise. 


%





\bibliographystyle{IEEEtran}
\bibliography{IEEEabrv}

\begin{thebibliography}{10}
\providecommand{\url}[1]{#1}
\csname url@samestyle\endcsname
\providecommand{\newblock}{\relax}
\providecommand{\bibinfo}[2]{#2}
\providecommand{\BIBentrySTDinterwordspacing}{\spaceskip=0pt\relax}
\providecommand{\BIBentryALTinterwordstretchfactor}{4}
\providecommand{\BIBentryALTinterwordspacing}{\spaceskip=\fontdimen2\font plus
\BIBentryALTinterwordstretchfactor\fontdimen3\font minus
  \fontdimen4\font\relax}
\providecommand{\BIBforeignlanguage}[2]{{%
\expandafter\ifx\csname l@#1\endcsname\relax
\typeout{** WARNING: IEEEtran.bst: No hyphenation pattern has been}%
\typeout{** loaded for the language `#1'. Using the pattern for}%
\typeout{** the default language instead.}%
\else
\language=\csname l@#1\endcsname
\fi
#2}}
\providecommand{\BIBdecl}{\relax}
\BIBdecl

\bibitem{deng2009imagenet}
J.~Deng, W.~Dong, R.~Socher, L.-J. Li, K.~Li, and L.~Fei-Fei, ``Imagenet: A
  large-scale hierarchical image database,'' in \emph{2009 IEEE conference on
  computer vision and pattern recognition}.\hskip 1em plus 0.5em minus
  0.4em\relax Ieee, 2009, pp. 248--255.

\bibitem{li2020gradient}
M.~Li, M.~Soltanolkotabi, and S.~Oymak, ``Gradient descent with early stopping
  is provably robust to label noise for overparameterized neural networks,'' in
  \emph{International Conference on Artificial Intelligence and
  Statistics}.\hskip 1em plus 0.5em minus 0.4em\relax PMLR, 2020, pp.
  4313--4324.

\bibitem{zhang2018understanding}
C.~Zhang, S.~Bengio, M.~Hardt, B.~Recht, and O.~Vinyals, ``Understanding deep
  learning requires rethinking generalization,'' 2018.

\bibitem{zhu2004class}
X.~Zhu and X.~Wu, ``Class noise vs. attribute noise: A quantitative study,''
  \emph{Artificial intelligence review}, vol.~22, no.~3, pp. 177--210, 2004.

\bibitem{thulasidasan2019combating}
S.~Thulasidasan, T.~Bhattacharya, J.~Bilmes, G.~Chennupati, and J.~Mohd-Yusof,
  ``Combating label noise in deep learning using abstention,'' in
  \emph{International Conference on Machine Learning}.\hskip 1em plus 0.5em
  minus 0.4em\relax PMLR, 2019, pp. 6234--6243.

\bibitem{kim2019nlnl}
Y.~Kim, J.~Yim, J.~Yun, and J.~Kim, ``Nlnl: Negative learning for noisy
  labels,'' in \emph{Proceedings of the IEEE International Conference on
  Computer Vision}, 2019, pp. 101--110.

\bibitem{huang2019o2u}
J.~Huang, L.~Qu, R.~Jia, and B.~Zhao, ``O2u-net: A simple noisy label detection
  approach for deep neural networks,'' in \emph{Proceedings of the IEEE
  International Conference on Computer Vision}, 2019, pp. 3326--3334.

\bibitem{lu2021mixnn}
Y.~Lu and W.~He, ``Mixnn: Combating noisy labels in deep learning by mixing
  with nearest neighbors,'' in \emph{2021 IEEE International Conference on Big
  Data (Big Data)}.\hskip 1em plus 0.5em minus 0.4em\relax IEEE, 2021, pp.
  847--856.

\bibitem{arpit2017closer}
D.~Arpit, S.~Jastrzebski, N.~Ballas, D.~Krueger, E.~Bengio, M.~S. Kanwal,
  T.~Maharaj, A.~Fischer, A.~Courville, Y.~Bengio \emph{et~al.}, ``A closer
  look at memorization in deep networks,'' in \emph{Proceedings of the 34th
  International Conference on Machine Learning-Volume 70}.\hskip 1em plus 0.5em
  minus 0.4em\relax JMLR. org, 2017, pp. 233--242.

\bibitem{permuter2006study}
H.~Permuter, J.~Francos, and I.~Jermyn, ``A study of gaussian mixture models of
  color and texture features for image classification and segmentation,''
  \emph{Pattern recognition}, vol.~39, no.~4, pp. 695--706, 2006.

\bibitem{ghosh2017robust}
A.~Ghosh, H.~Kumar, and P.~Sastry, ``Robust loss functions under label noise
  for deep neural networks,'' in \emph{Thirty-First AAAI Conference on
  Artificial Intelligence}, 2017.

\bibitem{zhang2018generalized}
Z.~Zhang and M.~Sabuncu, ``Generalized cross entropy loss for training deep
  neural networks with noisy labels,'' in \emph{Advances in neural information
  processing systems}, 2018, pp. 8778--8788.

\bibitem{wang2019symmetric}
Y.~Wang, X.~Ma, Z.~Chen, Y.~Luo, J.~Yi, and J.~Bailey, ``Symmetric cross
  entropy for robust learning with noisy labels,'' in \emph{Proceedings of the
  IEEE/CVF International Conference on Computer Vision}, 2019, pp. 322--330.

\bibitem{ma2020normalized}
X.~Ma, H.~Huang, Y.~Wang, S.~Romano, S.~Erfani, and J.~Bailey, ``Normalized
  loss functions for deep learning with noisy labels,'' in \emph{International
  Conference on Machine Learning}.\hskip 1em plus 0.5em minus 0.4em\relax PMLR,
  2020, pp. 6543--6553.

\bibitem{tanaka2018joint}
D.~Tanaka, D.~Ikami, T.~Yamasaki, and K.~Aizawa, ``Joint optimization framework
  for learning with noisy labels,'' in \emph{Proceedings of the IEEE Conference
  on Computer Vision and Pattern Recognition}, 2018, pp. 5552--5560.

\bibitem{wang2018iterative}
Y.~Wang, W.~Liu, X.~Ma, J.~Bailey, H.~Zha, L.~Song, and S.-T. Xia, ``Iterative
  learning with open-set noisy labels,'' in \emph{Proceedings of the IEEE
  Conference on Computer Vision and Pattern Recognition}, 2018, pp. 8688--8696.

\bibitem{yi2019probabilistic}
K.~Yi and J.~Wu, ``Probabilistic end-to-end noise correction for learning with
  noisy labels,'' in \emph{Proceedings of the IEEE Conference on Computer
  Vision and Pattern Recognition}, 2019, pp. 7017--7025.

\bibitem{patrini2017making}
G.~Patrini, A.~Rozza, A.~Krishna~Menon, R.~Nock, and L.~Qu, ``Making deep
  neural networks robust to label noise: A loss correction approach,'' in
  \emph{Proceedings of the IEEE Conference on Computer Vision and Pattern
  Recognition}, 2017, pp. 1944--1952.

\bibitem{jiang2017mentornet}
L.~Jiang, Z.~Zhou, T.~Leung, L.-J. Li, and L.~Fei-Fei, ``Mentornet: Learning
  data-driven curriculum for very deep neural networks on corrupted labels,''
  in \emph{International Conference on Machine Learning}.\hskip 1em plus 0.5em
  minus 0.4em\relax PMLR, 2018, pp. 2304--2313.

\bibitem{malach2017decoupling}
E.~Malach and S.~Shalev-Shwartz, ``Decoupling" when to update" from" how to
  update",'' in \emph{Advances in Neural Information Processing Systems}, 2017,
  pp. 960--970.

\bibitem{han2018co}
B.~Han, Q.~Yao, X.~Yu, G.~Niu, M.~Xu, W.~Hu, I.~Tsang, and M.~Sugiyama,
  ``Co-teaching: Robust training of deep neural networks with extremely noisy
  labels,'' in \emph{Advances in neural information processing systems}, 2018,
  pp. 8527--8537.

\bibitem{lu2022ensemble}
Y.~Lu, Y.~Bo, and W.~He, ``An ensemble model for combating label noise,'' in
  \emph{Proceedings of the Fifteenth ACM International Conference on Web Search
  and Data Mining}, 2022, pp. 608--617.

\bibitem{li2020dividemix}
J.~Li, R.~Socher, and S.~C. Hoi, ``Dividemix: Learning with noisy labels as
  semi-supervised learning,'' \emph{ICLR}, 2020.

\bibitem{nguyen2020self}
T.~Nguyen, C.~Mummadi, T.~Ngo, L.~Beggel, and T.~Brox, ``Self: learning to
  filter noisy labels with self-ensembling,'' in \emph{International Conference
  on Learning Representations (ICLR)}, 2020.

\bibitem{li2019learning}
J.~Li, Y.~Wong, Q.~Zhao, and M.~S. Kankanhalli, ``Learning to learn from noisy
  labeled data,'' in \emph{Proceedings of the IEEE Conference on Computer
  Vision and Pattern Recognition}, 2019, pp. 5051--5059.

\bibitem{berthelot2019mixmatch}
D.~Berthelot, N.~Carlini, I.~Goodfellow, N.~Papernot, A.~Oliver, and C.~A.
  Raffel, ``Mixmatch: A holistic approach to semi-supervised learning,'' in
  \emph{Advances in Neural Information Processing Systems}, 2019, pp.
  5050--5060.

\bibitem{liu2020early}
S.~Liu, J.~Niles-Weed, N.~Razavian, and C.~Fernandez-Granda, ``Early-learning
  regularization prevents memorization of noisy labels,'' \emph{Advances in
  Neural Information Processing Systems}, vol.~33, 2020.

\bibitem{lu2021confidence}
Y.~Lu, Y.~Bo, and W.~He, ``Confidence adaptive regularization for deep learning
  with noisy labels,'' \emph{arXiv preprint arXiv:2108.08212}, 2021.

\bibitem{he2016deep}
K.~He, X.~Zhang, S.~Ren, and J.~Sun, ``Deep residual learning for image
  recognition,'' in \emph{Proceedings of the IEEE conference on computer vision
  and pattern recognition}, 2016, pp. 770--778.

\bibitem{van2008visualizing}
L.~Van~der Maaten and G.~Hinton, ``Visualizing data using t-sne.''
  \emph{Journal of machine learning research}, vol.~9, no.~11, 2008.

\bibitem{har2012approximate}
S.~Har-Peled, P.~Indyk, and R.~Motwani, ``Approximate nearest neighbor: Towards
  removing the curse of dimensionality,'' \emph{Theory of computing}, vol.~8,
  no.~1, pp. 321--350, 2012.

\bibitem{bentley1975multidimensional}
J.~L. Bentley, ``Multidimensional binary search trees used for associative
  searching,'' \emph{Communications of the ACM}, vol.~18, no.~9, pp. 509--517,
  1975.

\bibitem{malkov2018efficient}
Y.~A. Malkov and D.~A. Yashunin, ``Efficient and robust approximate nearest
  neighbor search using hierarchical navigable small world graphs,'' \emph{IEEE
  transactions on pattern analysis and machine intelligence}, vol.~42, no.~4,
  pp. 824--836, 2018.

\bibitem{zhang2017mixup}
H.~Zhang, M.~Cisse, Y.~N. Dauphin, and D.~Lopez-Paz, ``mixup: Beyond empirical
  risk minimization,'' \emph{arXiv preprint arXiv:1710.09412}, 2017.

\bibitem{thulasidasan2019mixup}
S.~Thulasidasan, G.~Chennupati, J.~A. Bilmes, T.~Bhattacharya, and S.~Michalak,
  ``On mixup training: Improved calibration and predictive uncertainty for deep
  neural networks,'' \emph{Advances in Neural Information Processing Systems},
  vol.~32, 2019.

\bibitem{szegedy2016rethinking}
C.~Szegedy, V.~Vanhoucke, S.~Ioffe, J.~Shlens, and Z.~Wojna, ``Rethinking the
  inception architecture for computer vision,'' in \emph{Proceedings of the
  IEEE conference on computer vision and pattern recognition}, 2016, pp.
  2818--2826.

\bibitem{krizhevsky2012imat}
A.~Krizhevsky, I.~Sutskever, and G.~E. Hinton, ``Imagenet classification with
  deep convolutional neural networks,'' in \emph{Advances in neural information
  processing systems}, 2012, pp. 1097--1105.

\bibitem{loshchilov2016sgdr}
I.~Loshchilov and F.~Hutter, ``Sgdr: Stochastic gradient descent with warm
  restarts,'' \emph{arXiv preprint arXiv:1608.03983}, 2016.

\bibitem{lu2022selc}
Y.~Lu and W.~He, ``Selc: Self-ensemble label correction improves learning with
  noisy labels,'' \emph{arXiv preprint arXiv:2205.01156}, 2022.

\bibitem{reed2014training}
S.~E. Reed, H.~Lee, D.~Anguelov, C.~Szegedy, D.~Erhan, and A.~Rabinovich,
  ``Training deep neural networks on noisy labels with bootstrapping,'' in
  \emph{ICLR (Workshop)}, 2015.

\bibitem{lee2019robust}
K.~Lee, S.~Yun, K.~Lee, H.~Lee, B.~Li, and J.~Shin, ``Robust inference via
  generative classifiers for handling noisy labels,'' in \emph{International
  Conference on Machine Learning}.\hskip 1em plus 0.5em minus 0.4em\relax PMLR,
  2019, pp. 3763--3772.

\bibitem{pmlr-v97-arazo19a}
E.~Arazo, D.~Ortego, P.~Albert, N.~O'Connor, and K.~Mcguinness, ``Unsupervised
  label noise modeling and loss correction,'' in \emph{Proceedings of the 36th
  International Conference on Machine Learning}, 2019, pp. 312--321.

\bibitem{xiao2015learning}
T.~Xiao, T.~Xia, Y.~Yang, C.~Huang, and X.~Wang, ``Learning from massive noisy
  labeled data for image classification,'' in \emph{Proceedings of the IEEE
  conference on computer vision and pattern recognition}, 2015, pp. 2691--2699.

\bibitem{li2017webvision}
W.~Li, L.~Wang, W.~Li, E.~Agustsson, and L.~V. Gool, ``Webvision database:
  Visual learning and understanding from web data.'' \emph{CoRR}, 2017.

\bibitem{chen2019understanding}
P.~Chen, B.~B. Liao, G.~Chen, and S.~Zhang, ``Understanding and utilizing deep
  neural networks trained with noisy labels,'' in \emph{International
  Conference on Machine Learning}, 2019, pp. 1062--1070.

\bibitem{szegedy2017inception}
C.~Szegedy, S.~Ioffe, V.~Vanhoucke, and A.~A. Alemi, ``Inception-v4,
  inception-resnet and the impact of residual connections on learning,'' in
  \emph{Thirty-first AAAI conference on artificial intelligence}, 2017.

\bibitem{chen2021beyond}
P.~Chen, J.~Ye, G.~Chen, J.~Zhao, and P.-A. Heng, ``Beyond class-conditional
  assumption: A primary attempt to combat instance-dependent label noise,'' in
  \emph{Proceedings of the AAAI Conference on Artificial Intelligence},
  vol.~35, no.~13, 2021, pp. 11\,442--11\,450.

\bibitem{zheng2020error}
S.~Zheng, P.~Wu, A.~Goswami, M.~Goswami, D.~Metaxas, and C.~Chen,
  ``Error-bounded correction of noisy labels,'' in \emph{International
  Conference on Machine Learning}.\hskip 1em plus 0.5em minus 0.4em\relax PMLR,
  2020, pp. 11\,447--11\,457.

\bibitem{ma2018dimensionality}
X.~Ma, Y.~Wang, M.~E. Houle, S.~Zhou, S.~M. Erfani, S.-T. Xia, S.~Wijewickrema,
  and J.~Bailey, ``Dimensionality-driven learning with noisy labels,''
  \emph{arXiv preprint arXiv:1806.02612}, 2018.

\bibitem{mirzasoleiman2020coresets}
B.~Mirzasoleiman, K.~Cao, and J.~Leskovec, ``Coresets for robust training of
  neural networks against noisy labels,'' \emph{Neural Information Processing
  Systems (NeurIPS)}, 2020.

\bibitem{song2020robust}
J.~Song, Y.~Dauphin, M.~Auli, and T.~Ma, ``Robust and on-the-fly dataset
  denoising for image classification,'' in \emph{European Conference on
  Computer Vision}.\hskip 1em plus 0.5em minus 0.4em\relax Springer, 2020, pp.
  556--572.

\end{thebibliography}
%



%

\begin{IEEEbiography}[{\includegraphics[width=1in,height=1.25in,clip,keepaspectratio]{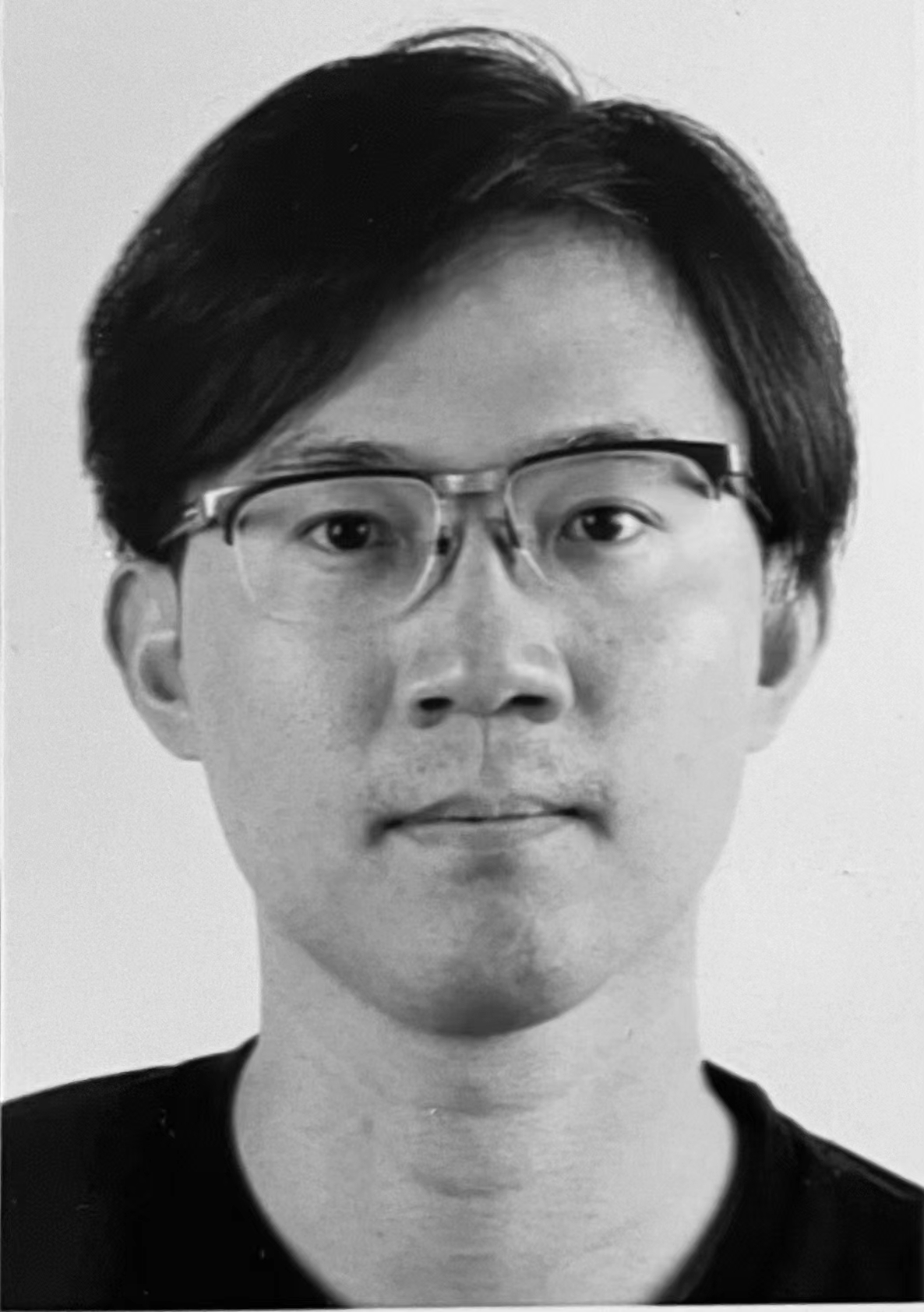}}]{Yangdi Lu}
	received the BSc degree from the Department of Computer Science, Sichuan University, Chengdu, China, in 2015. He received MSc degree from the Department of Computing and Software, McMaster University, in 2018. He is currently a PhD candidate with the Department of Computing and Software, McMaster University. His research interests include trustworthy machine learning, computer vision and information retrieval.
\end{IEEEbiography}
\vspace{-2em}
\begin{IEEEbiography}[{\includegraphics[width=1in,height=1.25in,clip,keepaspectratio]{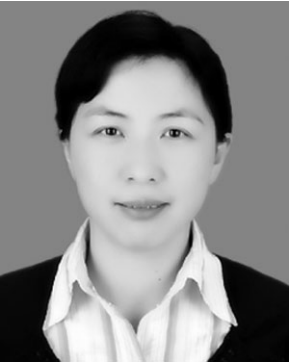}}]{Wenbo He}
received the PhD degree from the Department of Computer Science, University of Illinois at Urbana-Champaign, in 2008. She is currently an associate professor with the Department
of Computing and Software, McMaster University, and adjunct professor with the School of Computer Science, McGill University. Her research focuses on trustworthy machine learning, multimedia big data, cloud computing, information security and privacy, and pervasive computing, etc.
\end{IEEEbiography}







\end{document}


%
\title{Supplementary Materials}
%
%
%
%

\author{Yangdi Lu,
        Wenbo He,
\IEEEcompsocitemizethanks{\IEEEcompsocthanksitem Y. Lu and W. He are with the Department of Computing and Software, McMaster University, Hamilton, Ontario L8S 4L7, Canada.\protect\\
E-mail: luy100@mcmaster.ca, hew11@mcmaster.ca }
\thanks{Manuscript received May 6, 2022.}}

\maketitle

\section{Baselines}
\noindent\textbf{Baselines} We compare our method to the following baselines from different categories. (1) CE directly uses the standard cross-entropy loss to train the DNNs on noisy training data. (2) F-correction and Bootstrap belong to loss correction category. (3) GCE, SCE, NFL+MAE, and NCE+RCE belong to robust loss function category. (4) Joint Optim, PENCIL, RoG+D2L, M-correction, SEAL and LRT belong to label correction category. (4) Decoupling, Co-teaching, MentorNet and Iterative-CV belong to sample selection category. (4) O2U-net, NLNL, DAC, Crust and ODD belong to noisy pruning category. (5) SELF belongs to semi-supervised learning category. 

\section{Results on Correction Accuracy}

The results on other ratios of label noise on CIFAR-10 and CIFAR-100 are in Table \ref{table:correct_acc}.

\section{Confusion Matrix of Other Noise Ratios}

\begin{figure}[t]
	\centering
	\subfigure[40\% symm noise]{
		\includegraphics[width=0.65\columnwidth]{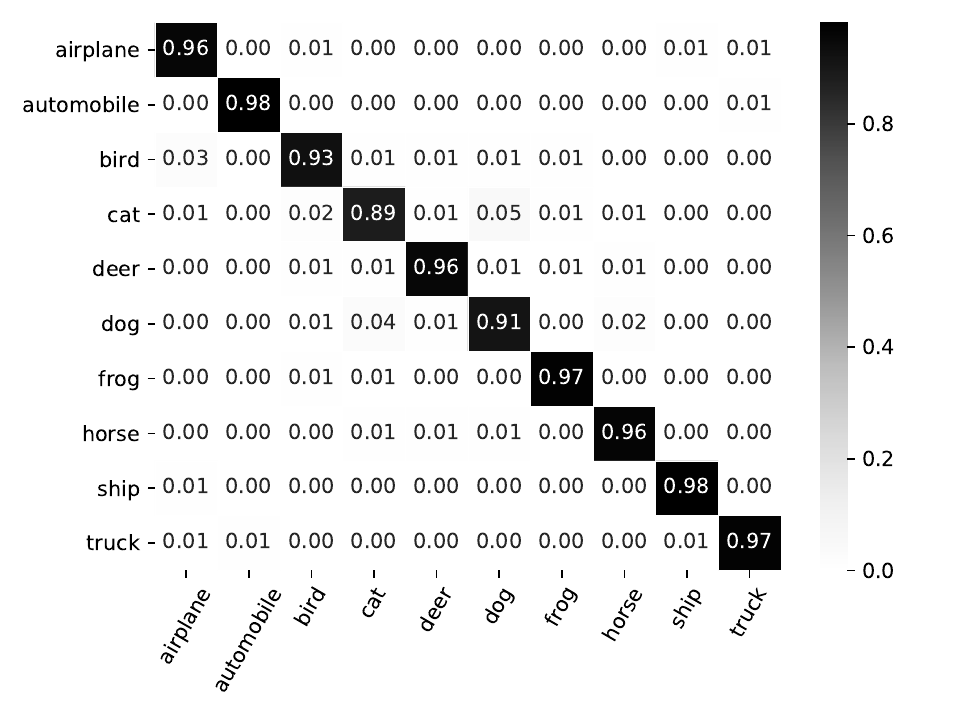}}
	\subfigure[80\% symm noise]{
		\includegraphics[width=0.65\columnwidth]{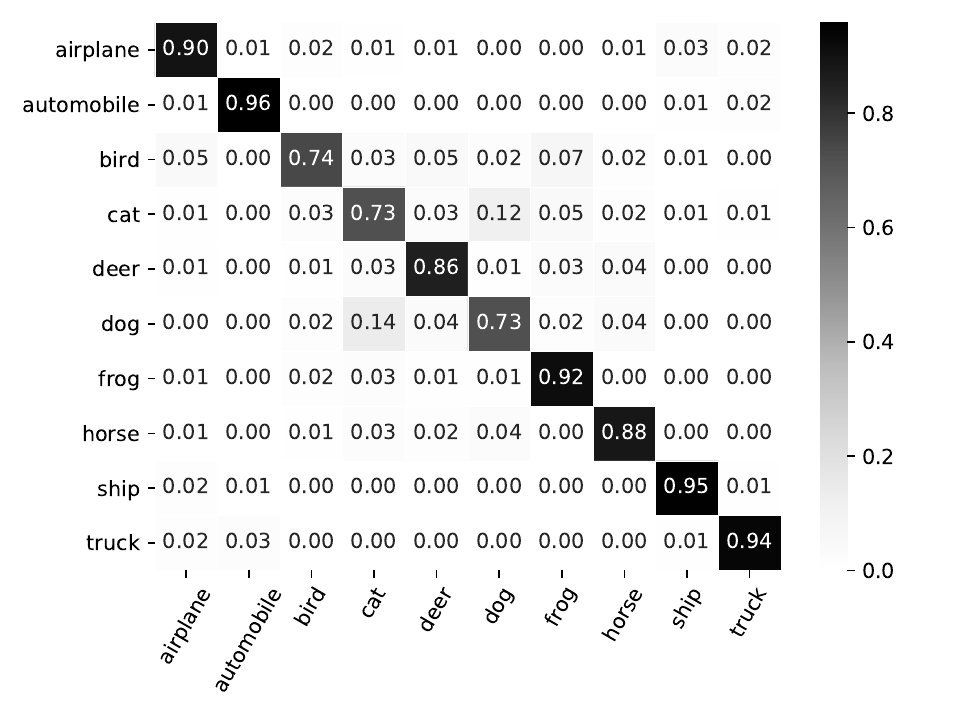}}
	\subfigure[40\% asymm noise]{
		\includegraphics[width=0.65\columnwidth]{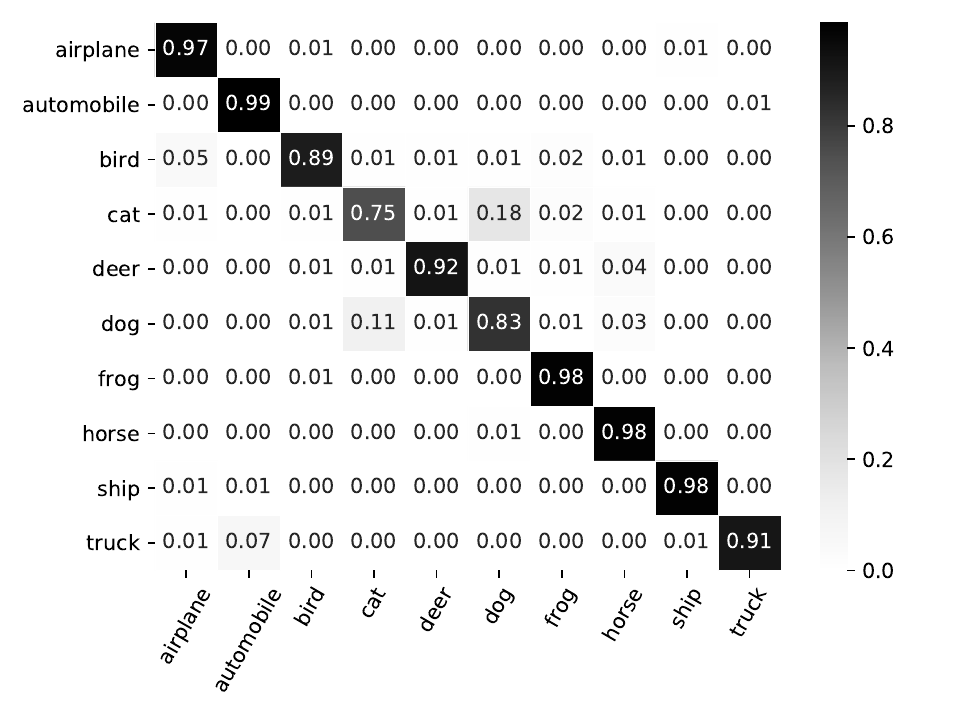}}
	\caption{Confusion matrix of corrected labels w.r.t clean labels on CIFAR-10 after MixNN.}
	\label{fig:label_correct}
\end{figure}

Figure \ref{fig:label_correct} shows the confusion matrix on other ratios of label noise.

\begin{table}[t]
	\begin{center}
		
		\caption{Correction accuracy (\%) on CIFAR-10 and CIFAR-100 with various levels of label noise injected to training set.}
		\label{table:correct_acc}
		\resizebox{0.5\textwidth}{!}{
			\centering
			\begin{tabular}{| p{25mm} c| c c  c c c|c c c c c| } 
				\hline
				\multicolumn{2}{|c|}{\multirow{1}{*}{Dataset}} & \multicolumn{5}{c|}{CIFAR-10} &\multicolumn{5}{c|}{CIFAR-100}\\\hline
				\multicolumn{2}{|c|}{\multirow{1}{*}{Noise type}} & \multicolumn{4}{c}{symm} &\multicolumn{1}{c|}{asymm} & \multicolumn{4}{c}{symm}&\multicolumn{1}{c|}{asymm} \\ 
				\multicolumn{2}{|c|}{\multirow{1}{*}{Noise ratio}}& 20\% & \multicolumn{1}{c}{40\%}&\multicolumn{1}{c}{60\%}&\multicolumn{1}{c}{80\%}& \multicolumn{1}{c|}{40\%} & \multicolumn{1}{c}{20\%}&\multicolumn{1}{c}{40\%}&\multicolumn{1}{c}{60\%}&\multicolumn{1}{c}{80\%}&\multicolumn{1}{c|}{40\%}\\
				\hline
				\multirow{1}{*}{Correction accuracy (\%)} &  & 96.78	 &95.25&92.89&86.77&91.81&90.71&84.86&74.93&52.45&72.40\\ 
				\hline
			\end{tabular}
		}
	\end{center}
	
\end{table}